\documentclass[runningheads]{llncs}

\usepackage{eccv}

\usepackage{eccvabbrv}

\usepackage{graphicx}
\usepackage{booktabs}
\usepackage{multirow}
\usepackage{pifont}
\usepackage{xcolor,colortbl}
\usepackage{pifont}
\usepackage{tikz}
\usepackage{twemojis}
\usepackage[normalem]{ulem}
\usepackage{algorithm}
\usepackage{algorithmic}
\usepackage{amsmath}
\usepackage{amssymb}
\usepackage{bm}
\usepackage{array}
\usepackage[accsupp]{axessibility}  

\usepackage{orcidlink}

\newcommand{\para}[1]{\vspace{.05in}\noindent\textbf{#1}\quad}

\newcommand{\cmark}{\ding{51}}
\newcommand{\xmark}{\ding{55}}

\newcommand{\good}[1]{{\color{ForestGreen}#1}}
\newcommand{\bad}[1]{{\color{red!70}#1}}

\begin{document}

\title{\texorpdfstring{\scalebox{1.8}{\twemoji{books}}}{} STAC: Selective Spatiotemporal Aggregation and Compression for Video Reasoning Segmentation}

\titlerunning{STAC}


\author{%
Syed Ariff Syed Hesham\textsuperscript{1,2}, 
Yun Liu\textsuperscript{3,4,5}\thanks{Corresponding author: Yun Liu (liuyun@nankai.edu.cn)}, 
Guolei Sun\textsuperscript{3,4,5}, 
Jing Yang\textsuperscript{6}, \\ 
Henghui Ding\textsuperscript{7}, 
Xue Geng\textsuperscript{2}, 
Xudong Jiang\textsuperscript{1}  \\
}

\authorrunning{~Hesham et al.}

\institute{School of EEE, Nanyang Technological University, Singapore \and
Institute for Infocomm Research, A*STAR, Singapore
 \and
VCIP, CS, Nankai University, Tianjin, China \and
AAIS, Nankai University, Tianjin, China \and
NKIARI, Shenzhen Futian, China \and
Guizhou University, Guizhou, China \and
Fudan University, Shanghai, China
}

\maketitle

\begin{abstract}
Video reasoning segmentation demands pixel-accurate object tracking across hundreds of frames under complex natural language queries, producing dense spatiotemporal tokens whose quadratic self-attention cost makes long-video processing prohibitive. Existing methods address this through token compression, yet typically operate on encoder features lacking temporal context, constraining selection before content redundancy can be reliably assessed. Informed compression requires contextual awareness, but acquiring that awareness at full resolution incurs the same quadratic cost compression aims to reduce. State-space models resolve this constraint, as their linear recurrence selectively conditions each token on temporal context at $\mathcal{O}(T)$ cost, producing representations where content redundancy becomes assessable. Building on this, \textbf{S}elective Spatio\textbf{T}emporal \textbf{A}ggregation and \textbf{C}ompression (\textbf{STAC}) enriches features via decoupled bidirectional spatial and causal temporal scanning, leveraging recurrence-derived redundancy for hierarchical compression with adaptive thresholds optimised with segmentation objective. STAC achieves 85\% token reduction and 1.8$\times$ speedup while surpassing compression-free baselines on reasoning segmentation benchmarks in a zero-shot streaming-compatible setting. Code is available \href{https://github.com/MCG-NKU/nku-video}{here}. 
\end{abstract}

\section{Introduction}
\label{sec:intro}
\vspace{-0.4em}

Large Language Models (LLMs)~\cite{achiam2023gpt,touvron2023llama2,dubey2024llama,jiang2024mixtral} have transformed the field of artificial intelligence, demonstrating remarkable capabilities in reasoning and task completion. Building upon this foundation, Multimodal Large Language Models (MLLMs) have extended these capabilities to visual understanding, achieving impressive results on image comprehension~\cite{bai2023qwenvl,liu2023visual,li2024llava,liu2024llavanext,chen2023internvl} and video analysis~\cite{cheng2024videollama,lin2023videollava,wang2024internvideo2,team2023gemini,zhang2023video,ning2026video}. Among these capabilities, video reasoning segmentation~\cite{lai2024lisa,bai2024one,yan2024visa} represents a particularly demanding challenge that requires models to interpret complex natural language queries, reason about spatiotemporal relationships, and produce pixel-accurate masks across extended sequences.

Most video MLLMs process videos by encoding each frame independently with pretrained image encoders~\cite{radford2021learning, zhai2023sigmoid}, concatenating the resulting tokens, and passing them to a language model~\cite{liu2023visual,liu2024llavanext,bai2023qwenvl}. This strategy proves effective for short clips spanning several seconds where token streams remain manageable and self-attention mechanisms can discover temporal relationships. However, this breaks down for longer videos due to quadratic attention complexity \cite{liu2024vision} and token explosion. For example, a 2-minute video at 2 FPS with a vision encoder~\cite{zhai2023sigmoid} producing 729 tokens per frame generates 61K tokens, nearly twice the typical 32K context window~\cite{dubey2024llama,jiang2024mixtral}, making long-video processing impractical~\cite{chen2024longvila}.

To address these challenges, most long-video methods adopt compression strategies that use different approaches to reduce the overall token sequence. Frame selection approaches~\cite{chen2024longvila,korbar2019scsampler,shen2024longvu,lin2025glus} target temporal redundancy but risk eliminating critical motion cues for tracking. An alternative strategy employs spatial compression~\cite{chen2024image,ye2024atpllava,tang2025tspo,song2024moviechat}, applying pooling or attention-based aggregation to reduce the number of per-frame tokens. However, operating within isolated frames limits access to cross-frame dependencies. Recent work~\cite{shen2024longvu,hu2025m,lin2023videollava,zhang2023video} combines both strategies, though compression decisions typically rely on predetermined heuristics or frame-level features rather than learned selection informed by global spatiotemporal context.

\begin{table}[t]
  \centering
  \scriptsize 
  \caption{Comparison of architectural characteristics across video compression approaches for reasoning segmentation. Brackets $(\checkmark)$ indicate partial support.}
  \label{tab:method_comparison}
  \begin{tabular}{@{}lcccc@{}}
    \toprule
    \textbf{Method} & \textbf{Temporal} & \textbf{Task-Aware} & \textbf{Hierarchical} & \textbf{Online} \\
    & \textbf{Integration} & \textbf{Compression} & \textbf{Understanding} & \textbf{Processing} \\
    \midrule
    \rowcolor{gray!15} \multicolumn{5}{l}{\textit{Local Compression Methods}} \\
    FastV~\cite{chen2024image}        & \bad{\xmark} & \bad{\xmark} & \bad{\xmark} & \bad{\xmark} \\
    ATP\,-LLaVA~\cite{ye2024atpllava} & \bad{\xmark} & \good{\cmark} & \bad{\xmark} & \bad{\xmark} \\
    TempMe~\cite{shen2024tempme}      & \good{\cmark} & (\good{\cmark}) & \good{\cmark} & \bad{\xmark} \\
    \midrule
    \rowcolor{gray!15} \multicolumn{5}{l}{\textit{Adaptive Compression Methods}} \\
    LongVU~\cite{shen2024longvu}   & \good{\cmark} & \good{\cmark} & \bad{\xmark} & \bad{\xmark} \\
    M\!-LLM~\cite{hu2025m}          & \good{\cmark} & \good{\cmark} & \bad{\xmark} & \bad{\xmark} \\
    TSPO~\cite{tang2025tspo}       & \good{\cmark} & \good{\cmark} & \bad{\xmark} & \bad{\xmark} \\
    \midrule
    \rowcolor{gray!15} \multicolumn{5}{l}{\textit{Bidirectional SSM Methods}} \\
    VideoMamba~\cite{li2024videomamba} & \good{\cmark} & \bad{\xmark} & \bad{\xmark} & \bad{\xmark} \\
    BIMBA~\cite{islam2025bimba}    & \good{\cmark} & (\good{\cmark}) & \bad{\xmark} & \bad{\xmark} \\
    STORM~\cite{jiang2025token}    & \good{\cmark} & \bad{\xmark} & \bad{\xmark} & \bad{\xmark} \\
    \midrule
    \rowcolor{green!15} \textbf{STAC (Ours)} & \good{\cmark} & \good{\cmark} & \good{\cmark} & \good{\cmark} \\
    \bottomrule
  \end{tabular}
  \vspace{-0.3ex}
\end{table}

These compression strategies, while effective, share a fundamental limitation. By operating on raw encoder features, they commit to token reduction before the model has understood the video's spatiotemporal structure. This creates a tension between compression and understanding, since acquiring contextual awareness at full resolution is precisely the cost compression aims to avoid. We observe that state-space models (SSMs)~\cite{gu2020hippo,gu2021combining,gu2022efficiently,gu2022s4d} resolve this tension. Their linear recurrence selectively conditions each hidden state on current and past content at $\mathcal{O}(T)$ cost, such that already-absorbed tokens produce near-identical enriched representations. This provides a redundancy signal intrinsic to the model's dynamics, a capability unique to persistent state accumulation.
Among SSMs, Mamba~\cite{gu2024mamba} introduces selective state-propagation mechanisms~\cite{gu2021efficiently} that dynamically retain relevant information while filtering redundancy. Extensions to the video domain~\cite{li2024videomamba,chen2024video,park2024videomamba,lu2024videomambapro} confirm these properties for temporal modelling, but most rely on unified bidirectional scanning~\cite{islam2025bimba,jiang2025token} that treats the spatiotemporal volume as a single block, hindering streaming deployment. A fundamental asymmetry underlies this limitation, as spatial relationships within a frame are non-causal with objects interacting with all neighbours simultaneously, whereas temporal evolution follows strict causal progression where future frames cannot influence past observations. This motivates decoupling spatial and temporal scanning to maintain both efficiency and streaming compatibility.

Building on this observation, \textbf{S}elective Spatio\textbf{T}emporal \textbf{A}ggregation and \textbf{C}ompression (\textbf{STAC}) exploits this SSM-derived redundancy signal for principled video token compression. Unlike existing SSM-based video methods that employ state-space model purely as a processing backbone, STAC enriches features through SSM before compressing them, ensuring redundancy decisions reflect tokens the recurrence has already absorbed rather than surface-level similarity. Specifically, our State-informed Spatiotemporal Aggregator (SSA) first enriches encoder features through bidirectional spatial and causal temporal scanning that respects each dimension's distinct structure, with the causal design independent of future observations to enable streaming-compatible deployment. Our Hierarchical State-adaptive Compression (HSC) module then leverages this enriched feature space for hierarchical temporal-then-spatial reduction, where adaptive thresholds respond to local content dynamics, compressing aggressively during static segments while preserving tokens at motion boundaries. Finally, task-grounded optimisation propagates segmentation gradients back into the compression policy, closing the loop so the downstream task directly determines which tokens are retained. This integrated design achieves $\sim$85\% token reduction with $\sim$1.8$\times$ speedup while surpassing compression-free baselines on reasoning segmentation benchmarks in a zero-shot, streaming-compatible setting.

Our primary contributions include:
\begin{itemize}

\item \textbf{State-informed Spatiotemporal Aggregator (SSA)} that enriches encoder features through decoupled bidirectional spatial and causal temporal scanning, producing a semantically grounded feature space where content redundancy becomes discernible before compression, while supporting streaming deployment through its causal temporal design.

\item \textbf{Hierarchical State-adaptive Compression (HSC)} that performs temporal -then-spatial token reduction in the recurrence-enriched feature space with adaptive thresholds responding to local content dynamics, compressing static regions aggressively while preserving motion boundaries.

\item \textbf{Task-Grounded Differentiable Compression} that propagates segmentation gradients directly through discrete compression decisions via straight-through estimation, aligning token retention with downstream mask accuracy and yielding content-adaptive policies with zero-shot transferability to unseen reasoning benchmarks.

\end{itemize}

\vspace{-0.8em}
\section{Related Work}
\label{sec:related}

\vspace{-0.3em}
\subsection{Language-Guided Video Segmentation}

Video reasoning segmentation requires interpreting complex natural-language queries to produce pixel-accurate masks while maintaining temporal consistency across extended sequences~\cite{xu2018youtubevos,ding2023mevis,lai2024lisa}. Traditional referring video object segmentation employs specialized transformers with explicit cross-modal fusion~\cite{gavrilyuk2018actor,khoreva2018video,botach2022endtoend}, with methods like ReferFormer~\cite{wu2022language} using deformable attention for multi-frame processing and MTTR~\cite{botach2022endtoend} introducing end-to-end temporal architectures. However, these lack flexible reasoning for complex queries requiring world knowledge or multi-hop inference~\cite{lin2025glus}. MLLM-based approaches integrate vision foundation models with large language models~\cite{kirillov2023segany,liu2023visual,lai2024lisa}, with LISA~\cite{lai2024lisa} pioneering embedding-as-mask paradigm for reasoning-based segmentation and video extensions introducing temporal modeling~\cite{yan2024visa,bai2024one,rasheed2024glamm}. VISA~\cite{yan2024visa} employs hierarchical encoding with temporal propagation, VideoLISA~\cite{bai2024one} proposes sparse-dense sampling with one-token-seg-all, and GLUS~\cite{lin2025glus} applies comprehensive global-local reasoning. Concurrently, VRS-HQ~\cite{gong2025devil} introduces hierarchical frame-level and temporal tokens with dynamic aggregation and occlusion-aware keyframe selection via SAM2, while Sa2VA~\cite{yuan2025sa2va} unifies SAM2 with LLaVA into a shared token space where LLM-generated instruction tokens guide mask prediction across images and videos. While these approaches are effective offline, they process video tokens uniformly without explicit compression~\cite{yan2024visa,bai2024one,lin2025glus}, treating all frames with equal computational weight and requiring access to complete sequences. This uniform treatment inherently limits scalability, prevents online deployment, and ignores the temporal redundancy that could potentially be exploited to enhance efficiency without sacrificing the fine-grained motion cues essential for reasoning.

\vspace{-0.5em}
\subsection{Efficient Video Understanding}

Conventional video understanding methods have extensively explored efficient spatiotemporal modeling to reduce redundant computation while preserving motion-sensitive representations \cite{karacan2025full,han2026object,ariff2026evaluating,an2026video,wang2026SAMDistill,feng2026training}. This concern becomes more pronounced in modern token-based video reasoning systems:
video MLLMs generate hundreds of tokens per frame with the use of pretrained encoders~\cite{radford2021learning,zhai2023sigmoid,liu2023visual}, rapidly exceeding context limits~\cite{dubey2024llama,jiang2024mixtral}. \textit{Token compression} mitigates spatial redundancy via learned selection~\cite{rao2021dynamicvit,pan2021ia} or attention-based pruning~\cite{bolya2022token,chen2024image,ye2024atpllava}. DynamicViT~\cite{rao2021dynamicvit} hierarchically prunes tokens, ToMe~\cite{bolya2022token} merges via bipartite matching, FastV~\cite{chen2024image} performs early pruning, and ATP-LLaVA~\cite{ye2024atpllava} introduces adaptive layer-wise thresholds. Video extensions exploit temporal redundancy~\cite{shen2024tempme,jiang2025token,song2024moviechat}, with TempMe~\cite{shen2024tempme} merging cross-frame tokens and STORM~\cite{jiang2025token} applying Mamba layers. \textit{Frame selection} reduces temporal dimensionality through uniform subsampling~\cite{chen2024longvila,lin2025glus} or adaptive scoring~\cite{korbar2019scsampler,shen2024longvu,hu2025m,tang2025tspo}. LongVU~\cite{shen2024longvu} leverages similarity-based filtering, while M-LLM~\cite{hu2025m} and TSPO~\cite{tang2025tspo} employ spatial pooling or reinforcement learning. However, these methods often rely on static ratios which are unsuited for varying complexity or heuristics lacking supervision, and universally require offline access to complete video sequences.

\subsection{State-Space Models for Visual Understanding}

State-space models building on classical theory~\cite{kalman1960new} provide efficient sequence processing through structured representations~\cite{gu2020hippo,gu2021combining,gu2022efficiently,gu2022s4d}. Mamba~\cite{gu2024mamba} introduces selective scan mechanisms that achieve linear complexity via input-dependent state propagation, retaining salient features while filtering redundancy~\cite{gu2021efficiently}. Vision applications adapt these principles through directional scanning~\cite{zhu2024vision,liu2024vmamba,pei2025efficientvmamba}, with Vision mamba~\cite{zhu2024vision} employing bidirectional scanning and VMamba~\cite{liu2024vmamba} introducing cross-scan mechanisms. Video extensions apply SSM across spatiotemporal volumes~\cite{li2024videomamba,chen2024video,park2024videomamba,lu2024videomambapro,yang2024vivim}, with VideoMamba~\cite{li2024videomamba} adopting bidirectional scanning and VideoMambaPro~\cite{lu2024videomambapro} addressing historical state decay through masked backward computation. Recent work integrates mamba into video MLLMs~\cite{islam2025bimba,jiang2025token,zhao2025cobra}, with BIMBA~\cite{islam2025bimba} introducing bi-directional spatiotemporal token selection with interleaved visual queries. However, existing architectures apply a unified bidirectional scan across complete sequences~\cite{li2024videomamba,park2024videomamba,islam2025bimba,jiang2025token}, treating spatial and temporal dimensions uniformly despite temporal evolution being strictly causal unlike symmetric spatial relationships. This requires full video access, preventing incremental frame-by-frame computation essential for online processing~\cite{islam2025bimba,shen2024tempme,wu2023onlinerefer}. Furthermore, prior work focuses on end-to-end processing~\cite{jiang2025token,islam2025bimba} rather than leveraging selective conditioning to enrich intermediate representations before task-specific selection in dense prediction. 

Our work addresses these gaps by compressing visual tokens \textit{upstream} through selective state conditioning, architecturally decoupling bidirectional spatial scanning from causal temporal scanning, and propagating segmentation gradients through compression decisions via task-grounded optimisation to enable efficient, streaming-compatible reasoning segmentation.

\section{Methodology}
\label{sec:methodology}

\begin{figure*}[!t]
    \centering
    \includegraphics[width=1\linewidth]{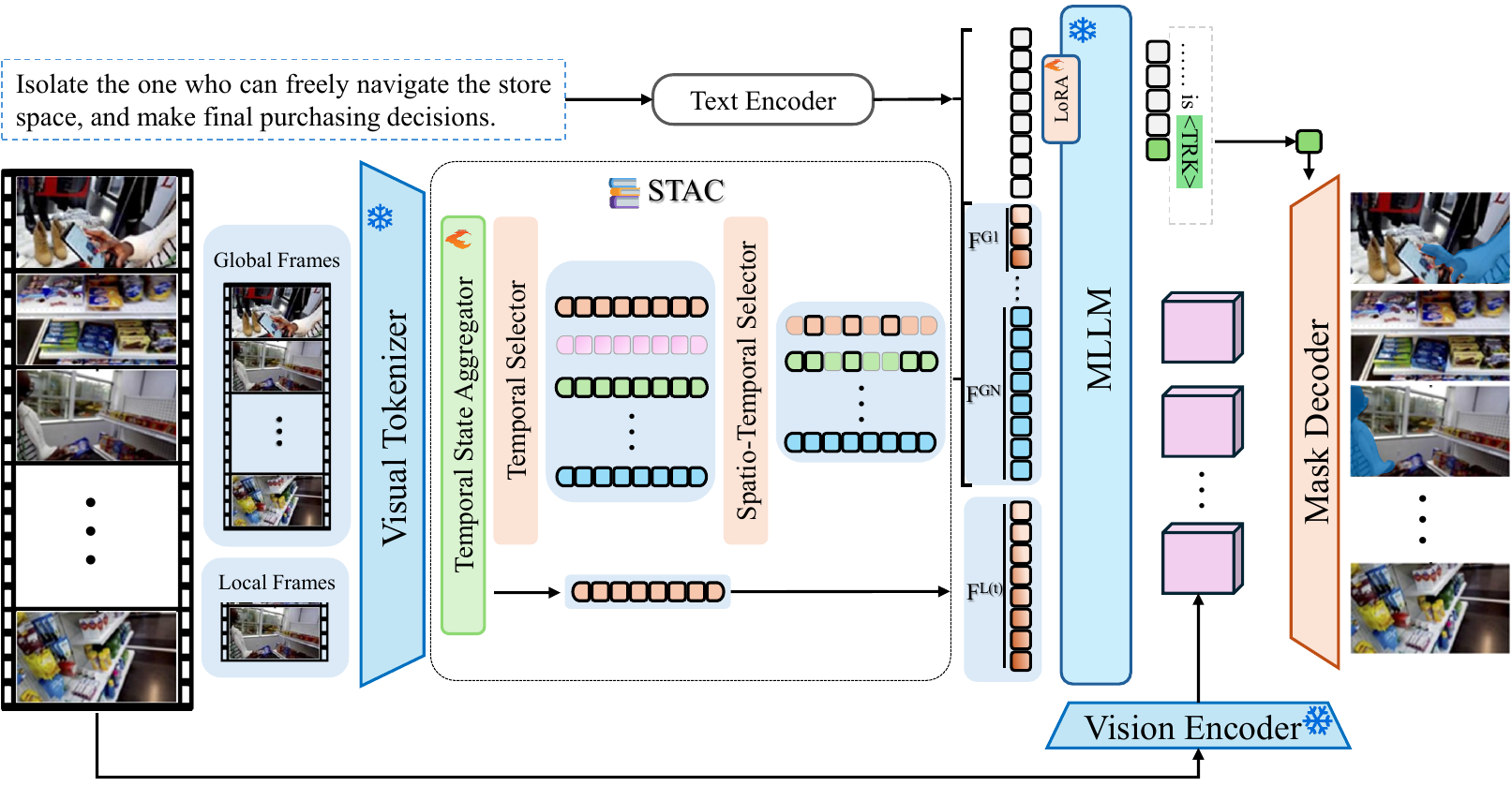}
    \caption{Our architecture introduces a generic spatiotemporal summarizer, which allows for temporal selection and spatiotemporal compression to computation while improving overall performance of the model.}
    \label{fig:architecture}
    \vspace{-1.5em}
\end{figure*}

\subsection{Architecture Overview}

Video reasoning segmentation requires discriminating temporally redundant content from semantically critical motion cues across long sequences. Since compression applied to raw encoder features lacks this temporal awareness, STAC places selective state-space conditioning before compression, ensuring token reduction operates on features where redundancy reflects integrated temporal context rather than raw appearance alone (\cref{fig:architecture}). Given video $\mathbf{V} \in \mathbb{R}^{T \times 3 \times H \times W}$ with frames $\{\mathbf{I}_1, \ldots, \mathbf{I}_T\}$ and text query $q$, we generate segmentation masks $\mathbb{M} = \{m_1, \ldots, m_T\}$ where $m_t \in \{0,1\}^{H \times W}$. Following standard protocols~\cite{bai2023qwenvl,liu2024llavanext}, we encode frames through frozen CLIP ViT-L/14~\cite{radford2021learning} to produce $N=256$ tokens per frame with features $\mathbf{F}_e \in \mathbb{R}^{T \times N \times d_e}$ where $d_e=1024$. The framework realises this through three cascaded stages, each respecting the distinct causal structure of spatial and temporal dimensions:

\vspace{-1em}
\begin{enumerate}
\item \textbf{Stage 1: Decoupled Spatiotemporal Scanning.} We transform independent frame tokens into globally-aware representations $\mathbf{F}_{\text{ST}} \in \mathbb{R}^{T \times N \times d_e}$ via a decoupled scanning architecture built on cascaded Mamba modules~\cite{gu2024mamba,dao2024transformers}. Diverging from architectures that treat spatiotemporal volumes as unified isotropic blocks, we systematically \textit{decouple} dimensions via bidirectional spatial scanning for semantic completeness and strictly causal temporal scanning for motion evolution. This separation guarantees $\mathcal{O}(TNd_e)$ linear complexity and prevents future-frame leakage to enable incremental inference.
\item \textbf{Stage 2: Asymmetric Causal Compression.} To circumvent latency bottlenecks in offline clustering, we implement an Asymmetric ``Predict-then-Compress'' policy where the current frame $\mathbf{F}_{\text{ST}}[t]$ is processed at full resolution for the immediate timestep, while all preceding frames are \textit{subsequently} compressed into a compact history $\mathbf{F}_c$ ($T' \ll T$) serving as temporal memory.

\item \textbf{Stage 3: Task-Objective Optimization.} We generate the $\langle\text{TRK}\rangle$ token for mask decoding~\cite{ravi2024sam} by projecting the compressed features $\mathbf{F}_c$ to dimension $D=4096$ while treating compression as a differentiable component. Instead of relying on generic heuristic reconstruction metrics, we train end-to-end to propagate segmentation gradients directly through discrete compression decisions via straight-through estimation~\cite{bengio2013estimating}, aligning the compression policy specifically to the segmentation objective.
\end{enumerate}
\vspace{-1em}

\begin{figure*}[!t]
    \centering
    \includegraphics[width=1\linewidth]{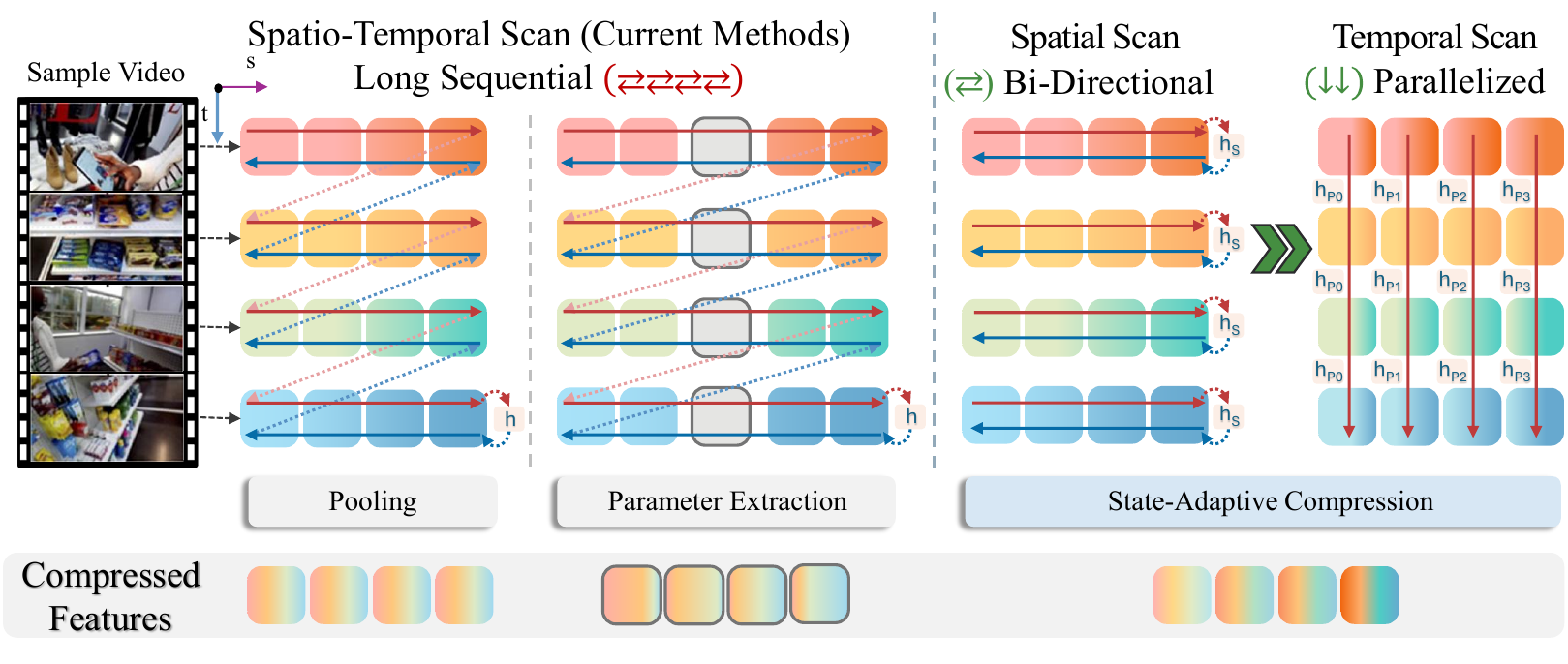}
    \caption{\textbf{Comparison of scanning strategies.} \textit{Current methods (left)} flatten videos into long T×H×W sequences requiring bidirectional processing over the entire video, and use fixed pooling~\cite{jiang2025token,li2024videomamba} or learnable parameter extraction~\cite{islam2025bimba} for compression. \textit{Our method (right)} decouples bidirectional spatial scanning from causal temporal scanning (spatially parallelized), enabling streaming-compatible content-adaptive compression via our HSC module while achieving superior token reduction. }
    \label{fig:spatiotemp}
    \vspace{-1.5em}
\end{figure*}

\subsection{State-informed Spatiotemporal Aggregator}
\label{ssec:SSA}

Standard visual encoders produce features $\mathbf{F}_e$ as isolated spatial patches devoid of global or temporal context. Inter-frame similarity on these representations captures surface-level visual resemblance rather than semantic redundancy, fundamentally limiting the quality of any compression decision derived from them. The proposed State-informed Spatiotemporal Aggregator (SSA) resolves this deficiency through a dual-stage architecture that enriches each token with spatiotemporal context \textit{before} compression. SSA realises this in two cascaded stages, integrating spatial context within each frame using bidirectional scanning, then propagating temporal context causally across frames to produce a feature space in which content redundancy becomes apparent.

\para{Spatial Aggregation.} Within a single frame, spatial relationships are non-causal with pixels interacting with all neighbors simultaneously. We capture this global context without imposing artificial ordering by applying bidirectional selective state scanning~\cite{zhu2024vision} independently across all $T$ frames. By averaging forward ($\mathcal{M}^b_{\rightarrow}$) and backward  ($\mathcal{M}^b_{\leftarrow}$) state scans via
\begin{equation}
\mathbf{F}_{\text{spatial}}[t] = \frac{1}{2}\left(\mathcal{M}^b_{\rightarrow}(\mathbf{F}_e[t]) + \mathcal{M}^b_{\leftarrow}(\mathbf{F}_e[t])\right),
\end{equation}
we effectively eliminate directional bias. This operation ensures the resulting features encode precise object boundaries and rich within-frame semantics which are essential prerequisites for pixel-level segmentation.

\para{Temporal Aggregation.} Conversely, temporal evolution adheres to strict physical causality. We respect this constraint to enable online streaming by applying causal selective state scanning $\mathcal{M}_{\text{causal}}$~\cite{dao2024transformers} efficiently parallelized across $N$ spatially independent locations. The mechanism accumulates temporal context through a recurrent update
\begin{equation}
\mathbf{F}_{\text{ST}}[n] = \mathcal{M}_{\text{causal}}(\mathbf{F}_{\text{spatial}}[:,n]),
\label{eq:2}
\end{equation}
where the causal selective state-space mechanism for location $n$ accumulates temporal context through
\begin{equation}
\mathbf{h}_{n,t} = \bar{\mathbf{A}}_t \mathbf{h}_{n,t-1} + \bar{\mathbf{B}}_t \mathbf{F}_{\text{spatial}}[t,n].
\label{eq:3}
\end{equation}

The input-dependent matrices $\bar{\mathbf{A}}_t, \bar{\mathbf{B}}_t \in \mathbb{R}^{d_e \times d_e}$ enable selective propagation of salient motion patterns while filtering redundancy. This causal structure supports incremental inference, as each incoming frame updates hidden states $\mathbf{h}_{n,t}$ without recomputing past representations, diverging from unified bidirectional methods~\cite{li2024videomamba,park2024videomamba,islam2025bimba} that require full video access and preclude streaming deployment. The resulting enriched features $\mathbf{F}_{\text{ST}}$ form the basis for the compression decisions that follow.

\vspace{-1em}

\subsection{Hierarchical State-adaptive Compression}
\label{ssec:STComp}

The proposed Hierarchical State-adaptive Compression (HSC) module performs hierarchical token reduction in the enriched feature space $\mathbf{F}_{\text{ST}}$, where the selective state conditioning of SSA (\cref{ssec:SSA}) has already integrated temporal context into each token. Within this space, consecutive frames whose content the recurrence has represented produce near-identical outputs, revealing redundancy that raw encoder features without such temporal integration cannot expose. HSC exploits this redundancy through a temporal-then-spatial hierarchy (\cref{tab:enrichment_timing}), first identifying and merging redundant frames before addressing within-frame spatial reduction, as premature spatial pooling would destroy the inter-frame differences this redundancy signal depends on.

\para{Adaptive Temporal Compression.}
Adjacent frames in video sequences exhibit significant temporal redundancy~\cite{korbar2019scsampler}, yet motion information resides precisely in the inter-frame differences that uniform downsampling discards. We identify redundancy via efficient $\mathcal{O}(T)$ sequential similarity. For each frame $t$, we compute frame-level representations via spatial averaging and measure similarity to the preceding frame:
\begin{equation}
\mathbf{g}_t = \frac{1}{N}\sum_{n=1}^N \mathbf{F}_{\text{ST}}[t,n], \quad s_t = \frac{\langle \mathbf{g}_t, \mathbf{g}_{t-1} \rangle}{\|\mathbf{g}_t\|_2 \|\mathbf{g}_{t-1}\|_2}.
\end{equation}

To circumvent the rigidity of fixed hyperparameters and ensure streaming compatibility, we avoid global statistics by estimating the distribution online via exponential moving averages (EMA). Let $\mu_t$ and $\sigma_t^2$ denote the running mean and variance at step $t$ with momentum $\alpha=0.1$:
\begin{equation}\label{eq:adaptiveth}
\mu_t = \alpha s_t + (1-\alpha)\mu_{t-1}, \quad \sigma_t^2 = \alpha (s_t - \mu_t)^2 + (1-\alpha)\sigma_{t-1}^2.
\end{equation}
The adaptive threshold is defined as $\tau_t = \mu_t + k \sigma_t$, where parameter $k$ is learned via a lightweight MLP. 
Consecutive frames satisfying $s_t > \tau_t$ are merged via averaging, producing compressed history $\mathbf{F}_{\text{temp}}$ with typical reduction of 60\%-70\% driven purely taking interframe content dynamics into account.

\para{Adaptive Spatial Compression.}
While temporal compression reduces frame count, spatial redundancy persists in background regions. We address this by measuring cross-frame coherence $c_n$ across the temporally-compressed sequence for each spatial location $n$ across the temporally-compressed sequence:

\begin{equation}
c_n = \frac{1}{T'-1} \sum_{t=2}^{T'} \frac{\langle \mathbf{F}_{\text{temp}}[t,n], \mathbf{F}_{\text{temp}}[t-1,n] \rangle}{\|\mathbf{F}_{\text{temp}}[t,n]\|_2 \|\mathbf{F}_{\text{temp}}[t-1,n]\|_2}.
\end{equation}
High coherence indicates temporally static patches whose representations across frames are suitable for merging, while low coherence highlights dynamic foreground regions requiring preservation. We compute adaptive thresholds using the statistic formulation in~\cref{eq:adaptiveth} applied per spatial location with local statistics $(\mu_c^{(n)}, \sigma_c^{(n)})$. Locations exceeding the threshold have their temporal tokens merged via averaging, while dynamic locations retain full temporal resolution. This process achieves 40\%-50\% further reduction concentrated on static regions, yielding compressed features $\mathbf{F}_c$ with $|\mathbf{F}_c| \ll T' \times N$ total tokens and a combined overall reduction of $\sim$85\%.

\vspace{-1em}
\subsection{Task-Grounded Differentiable Compression}
Standard compression metrics optimised for reconstruction may not guarantee preservation of segmentation-critical boundaries. The proposed framework instead integrates compression as a differentiable component within the segmentation pipeline, so that task gradients directly shape token retention through a joint objective
\begin{equation}
\mathcal{L}_{\text{total}} = \mathcal{L}_{\text{seg}} + \lambda \mathcal{L}_{\text{comp}},
\end{equation}
where $\mathcal{L}_{\text{seg}}$ comprises standard cross-entropy and Dice loss~\cite{ravi2024sam} with $\lambda=0.001$ balances the two objectives. The compression regularization term $\mathcal{L}_{\text{comp}}$ adapts to video complexity:
\begin{equation}
\mathcal{L}_{\text{comp}} = \frac{1}{B} \sum_{i=1}^B (1 - \hat{C}_i) \cdot \frac{c_i}{\sqrt{T_i}}.
\end{equation}
Here $B$ denotes batch size, $c_i$ counts retained tokens, $T_i$ denotes original token count, and $\hat{C}_i = \frac{1}{2}[(1-\mu_i) + \sigma_i]$ estimates complexity based on feature variance to reduce penalties for dynamic videos, and the square-root normalization $1/\sqrt{T_i}$ ensures sublinear scaling for long sequences. Such scaling incurs lower per-token penalties, preventing over-aggressive compression that would eliminate critical motion information in extended sequences while still incentivizing efficiency.

Alignment between compression and segmentation emerges through the learned thresholds in \cref{ssec:STComp}, which segmentation gradients adjust via the straight-through estimator~\cite{bengio2013estimating}. The threshold evolution (\cref{eq:adaptiveth}) stabilises these updates despite discrete selection, allowing training to converge on a content-adaptive policy that compresses aggressively during static segments while preserving tokens at motion boundaries. This content-adaptive policy completes the design loop in which decoupled scanning (\cref{ssec:SSA}) enriches features with spatiotemporal context, HSC translates the resulting redundancy into compression decisions, and the task objective calibrates those decisions to segmentation accuracy. As demonstrated in~\cref{fig:viscomp}, static videos sustain compression exceeding 90\% while action sequences retain 30\%--50\% of tokens for boundary precision.

\vspace{-1em}
\section{Experiments and Analysis}
\label{sec:experiments}
\vspace{-0.5em}

\subsection{Experimental Setup}
\label{ssec:exp_setup}

\noindent\textbf{Datasets.}\quad We evaluate on five benchmarks spanning two categories distinguished by query complexity. \textit{Referring benchmarks} require localizing objects through linguistic descriptions. Refer-YouTube-VOS~\cite{xu2018youtubevos} provides 3,978 videos with 15,458 expressions emphasizing appearance-based attributes and spatial relationships, while MeViS~\cite{ding2023mevis} comprises 2,006 videos with 28,570 expressions where targets must be disambiguated through motion patterns rather than static features. Ref-DAVIS17~\cite{khoreva2018video} extends DAVIS-2017~\cite{pont20172017} with 90 videos to evaluate performance under occlusion and appearance variation. \textit{Reasoning benchmarks} require multi-step inference integrating world knowledge. ReVOS~\cite{yan2024visa} provides 1,042 videos with 35,074 instruction-mask pairs demanding contextual inference beyond direct visual matching, while ReasonVOS~\cite{bai2024one} contains 91 videos with 458 samples specifically evaluating temporal understanding and causal reasoning across extended contexts. We report region similarity $\mathcal{J}$, contour accuracy $\mathcal{F}$, and their average $\mathcal{J}\&\mathcal{F}$ on official validation splits.

\para{Implementation.} Our architecture employs pretrained frozen CLIP ViT-L/14~\cite{radford2021learning} at 224$\times$224 resolution as the vision encoder, producing 256 tokens per frame with 1024-dimensional features. Features undergo spatiotemporal enrichment through our dual-stage SSA module maintaining 1024-dimensional hidden states, followed by HSC module for hierarchical compression via learned adaptive thresholds as described in~\cref{sec:methodology}. Compressed tokens are projected to 4096 dimensions and processed by LLaVA-7B~\cite{liu2023visual,touvron2023llama} with LoRA fine-tuning~\cite{hu2022lora} applied exclusively to attention projection matrices at rank 8 and $\alpha$=16, while base parameters remain frozen. SAM2~\cite{ravi2024sam} then decodes the language-aligned embeddings into segmentation masks. We train exclusively on referring segmentation datasets— MeViS~\cite{ding2023mevis} and Refer-YouTube-VOS~\cite{xu2018youtubevos}—withholding all reasoning segmentation data to evaluate zero-shot transfer on Ref-DAVIS17~\cite{khoreva2018video}, ReVOS~\cite{yan2024visa}, and ReasonVOS~\cite{bai2024one}. Training uses AdamW~\cite{adam2014method} optimizer with $\beta_1$=0.9, $\beta_2$=0.999, weight decay 0.05, learning rate $3\times10^{-4}$, and 100-step linear warmup. Each iteration processes batch size 2 with context frames sampled dynamically between 2-32 frames per video. Training completes 3,000 iterations in approximately 30 hours on 2$\times$ NVIDIA A40 GPUs.

\para{Baselines.} We compare against two categories isolating compression strategies. \textit{Compression-free baselines} include specialized referring architectures like ReferFormer~\cite{wu2022language}, OnlineRefer~\cite{wu2023onlinerefer}, and TempCD~\cite{tang2023temporal}, alongside MLLM-based methods LISA~\cite{lai2024lisa}, VISA~\cite{yan2024visa}, VideoLISA~\cite{bai2024one}, and GLUS~\cite{lin2025glus}, which process full token sequences without explicit reduction. \textit{Compression-based baselines} include 3D pooling with uniform spatiotemporal reduction, attention-based selection replacing Mamba modules with transformer layers and pruning via attention scores, Perceiver~\cite{jaegle2021perceiver} using learned cross-attention queries for aggregation, and BIMBA~\cite{islam2025bimba} employing bidirectional state-space compression. All baselines use official implementations with hyperparameters aligned to our configuration and equivalent reduction ratios for fair comparison.

\begin{table*}[t]
\centering
\caption{Comparison with state-of-the-art methods on referring and reasoning video object segmentation benchmarks. We report $\mathcal{J}$\&$\mathcal{F}$, $\mathcal{J}$, and $\mathcal{F}$ scores. Notably, STAC achieves competitive or superior performance across all benchmarks while operating on approximately $\sim$\textbf{15\% visual tokens} through adaptive compression, compared to other methods that process complete token sequences.}
\label{tab:main_results}
\resizebox{\textwidth}{!}{%
\begin{tabular}{l|ccc|ccc|ccc|ccc|ccc}
\toprule
\multirow{2}{*}{\textbf{Models}} & \multicolumn{3}{c|}{\textbf{Ref-DAVIS17}} & \multicolumn{3}{c|}{\textbf{MeViS}} & \multicolumn{3}{c|}{\textbf{Ref-YouTube}} & \multicolumn{3}{c|}{\textbf{ReasonVOS}} & \multicolumn{3}{c}{\textbf{ReVOS}} \\ \cmidrule{2-16}
& $\mathcal{J}$\&$\mathcal{F}$ & $\mathcal{J}$ & $\mathcal{F}$ & $\mathcal{J}$\&$\mathcal{F}$ & $\mathcal{J}$ & $\mathcal{F}$ & $\mathcal{J}$\&$\mathcal{F}$ & $\mathcal{J}$ & $\mathcal{F}$ & $\mathcal{J}$\&$\mathcal{F}$ & $\mathcal{J}$ & $\mathcal{F}$ & $\mathcal{J}$\&$\mathcal{F}$ & $\mathcal{J}$ & $\mathcal{F}$ \\
\midrule
\multicolumn{16}{l}{\textit{Referring Video Object Segmentation Methods}} \\
\midrule
URVOS~\cite{seo2020urvos} & 51.6 & 47.3 & 56.0 & 27.8 & 25.7 & 29.9 & 47.1 & 45.3 & 49.2 & - & - & - & - & - & - \\
RIS~\cite{ding2022language} & 54.3 & - & - & 29.3 & 27.8 & 30.8 & 49.4 & 48.2 & 50.6 & - & - & - & - & - & - \\
TempCD~\cite{tang2023temporal} & - & - & - & - & - & - & 65.8 & 63.6 & 68.0 & - & - & - & - & - & - \\
DsHmp~\cite{he2024decoupling} & 64.9 & 61.7 & 68.1 & 46.4 & 43.0 & 49.8 & 67.1 & 65.0 & 69.1 & - & - & - & - & - & - \\
OnlineRefer~\cite{wu2023onlinerefer} & 64.8 & 61.6 & 67.7 & - & - & - & 63.5 & 61.6 & 65.5 & 38.7 & 34.6 & 42.9 & - & - & - \\
SOC~\cite{luo2023soc} & - & - & - & - & - & - & 67.3 & 65.3 & 69.3 & 35.9 & 33.3 & 38.5 & - & - & - \\
SgMg~\cite{miao2023spectrum} & 63.3 & 60.6 & 66.0 & - & - & - & 65.7 & 63.9 & 67.4 & 36.2 & 33.7 & 38.7 & - & - & - \\
MTTR~\cite{botach2022endtoend} & - & - & - & 30.0 & 28.8 & 31.2 & 55.3 & 54.0 & 56.6 & 31.1 & 29.1 & 33.1 & 25.5& 25.1& 25.9\\
ReferFormer~\cite{wu2022language} & 61.1 & 58.1 & 64.1 & 31.0 & 29.8 & 32.2 & 62.9 & 61.3 & 64.6 & 32.9 & 30.2 & 35.6 & 28.1& 26.2& 29.9\\
LMPM~\cite{ding2023mevis} & 61.1 & 58.1 & 64.1 & 31.0 & 29.8 & 32.2 & 62.9 & 61.3 & 64.6 & 32.9 & 30.2 & 35.6 & 28.1& 26.2& 29.9\\
\midrule
\multicolumn{16}{l}{\textit{Multimodal LLM-based Reasoning Methods}} \\
\midrule
VideoGLaMM~\cite{munasinghe2025videoglamm} & 69.5& 65.6& 73.3& 45.2 & 42.1 & 48.2 & - & - & - & - & - & - & - & - & - \\
LISA-7B~\cite{lai2024lisa} & 64.8 & 62.2 & 67.3 & 37.2 & 35.1 & 39.4 & 53.9 & 53.4 & 54.3 & 31.1 & 29.1 & 33.1 & - & - & - \\
LISA-13B~\cite{lai2024lisa} & 66.0 & 63.2 & 68.8 & 37.9 & 35.8 & 40.0 & 54.4 & 54.0 & 54.8 & - & - & - & - & - & - \\
LLaMA-VID~\cite{li2024llama}+LMPM & - & - & - & - & - & - & - & - & - & - & - & - & 26.1& 20.9& 31.4\\
VideoLISA-3.8B (1-Tok)~\cite{bai2024one} & 67.7 & 63.8 & 71.5 & 42.3 & 39.4 & 45.2 & 61.7 & 60.2 & 63.3 & 45.1 & 43.1 & 47.1 & - & - & - \\
VideoLISA-3.8B (Post)~\cite{bai2024one} & 68.8 & 64.9 & 72.7 & 44.4 & 41.3 & 47.6 & 63.7 & 61.7 & 65.7 & 47.5 & 45.1 & 49.9 & - & - & - \\
TrackGPT-7B~\cite{zhu2023tracking} & 63.2 & 59.4 & 67.0 & 40.1 & 37.6 & 42.6 & 56.4 & 55.3 & 57.4 & - & - & - & 43.6& 41.8& 45.5\\
TrackGPT-13B~\cite{zhu2023tracking} & 66.5 & 62.7 & 70.4 & 41.2 & 39.2 & 43.1 & 59.5 & 58.1 & 60.8 & - & - & - & 45.0& 43.2& 46.8\\
VISA-Chat-7B~\cite{yan2024visa}& 69.4 & 66.3 & 72.5 & 43.5 & 40.7 & 46.3 & 61.5 & 59.8 & 63.2 & - & - & - & 46.1 & 43.9 & 48.2 \\
VISA-Chat-13B~\cite{yan2024visa}& 70.4 & 67.0 & 73.8 & 44.5 & 41.8 & 47.1 & 63.0 & 61.4 & 64.7 & - & - & - & 47.5 & 45.3 & 49.7 \\
GLUS$^{S}$~\cite{lin2025glus} & \uline{71.9} & \uline{68.3} & \uline{75.4} & \textbf{50.3} & \textbf{47.5} & \textbf{53.0} & \uline{64.8} & \uline{63.3} & \uline{66.4} & \uline{49.8} & \uline{47.2} & \uline{52.5} & \textbf{49.5} & \textbf{46.9} & \textbf{52.0} \\
\textbf{STAC (Ours)} & \textbf{73.6} & \textbf{70.1} & \textbf{77.2} & \uline{49.9} & \uline{47.2} & \uline{52.5} & \textbf{65.2} & \textbf{63.5} & \textbf{66.9} & \textbf{52.3} & \textbf{49.5} & \textbf{55.0} & \uline{48.4} & \uline{45.8} & \uline{51.3} \\
\bottomrule
\end{tabular}%

}
\end{table*}

\subsection{Main Results}

\cref{tab:main_results} compares STAC against state-of-the-art methods on referring and reasoning video object segmentation. Despite reducing visual tokens by 85\%, STAC achieves strong results on benchmarks emphasizing spatial localization and temporal reasoning. On Ref-DAVIS17~\cite{khoreva2018video} and Ref-YouTube~\cite{xu2018youtubevos}, our method surpasses all baseline compression-free approaches, demonstrating that context-aware selection enhances appearance-based referring segmentation even under occlusion. Most notably, despite training exclusively on referring segmentation data, STAC transfers effectively to reasoning benchmarks, outperforming the strongest compression-free baseline on ReasonVOS~\cite{bai2024one} in a zero-shot setting. This suggests that informed compression does not merely preserve reasoning ability but may actively improve it by forcing the model to retain only motion-critical content, filtering static redundancy that uncompressed approaches process.

On motion-centric benchmarks such as MeViS~\cite{ding2023mevis} and ReVOS~\cite{yan2024visa}, methods with full bidirectional temporal access achieve marginally stronger results. Yet STAC approaches comparable performance while compressing 85\% of visual tokens through selective state conditioning upstream before MLLM ingestion, with a causal temporal design that permits frame-by-frame online processing using only past observations. Furthermore, this upstream enrichment-before-compression strategy addresses the quadratic attention bottleneck before it reaches the MLLM, demonstrating that selective state conditioning captures sufficient temporal context from causal processing alone. This capability makes STAC distinctive among current approaches in combining competitive segmentation accuracy with substantial token compression and real-time deployability.

\begin{figure}[!t]
    \centering
    \begin{tikzpicture}
        \node[anchor=south west, inner sep=0] (main) at (0,0) {
            \includegraphics[width=0.63\columnwidth]{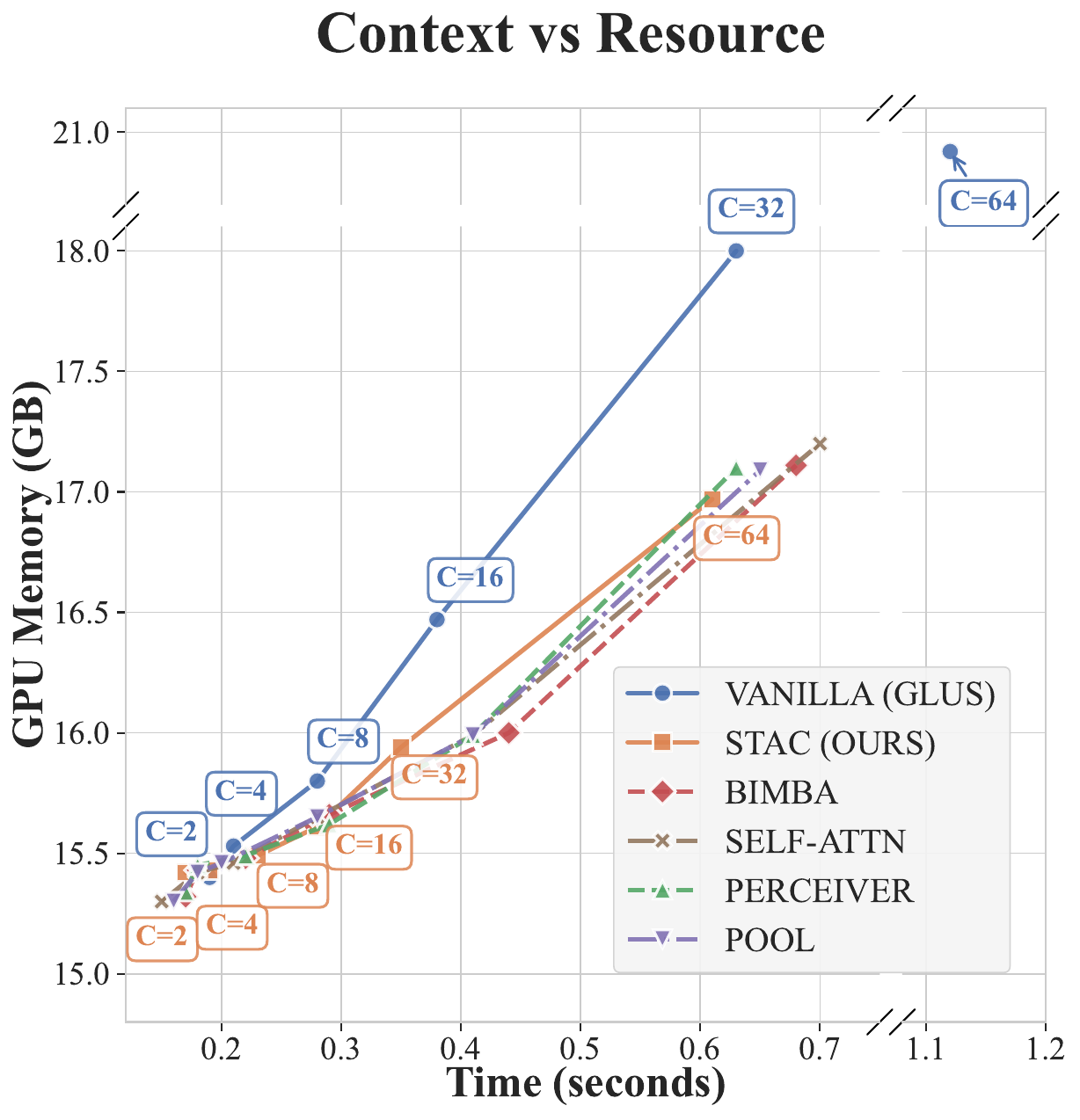}
        };
        \node[anchor=north west, inner sep=0] at ([xshift=28, yshift=-22]main.north west) {
            \includegraphics[width=0.251\columnwidth]{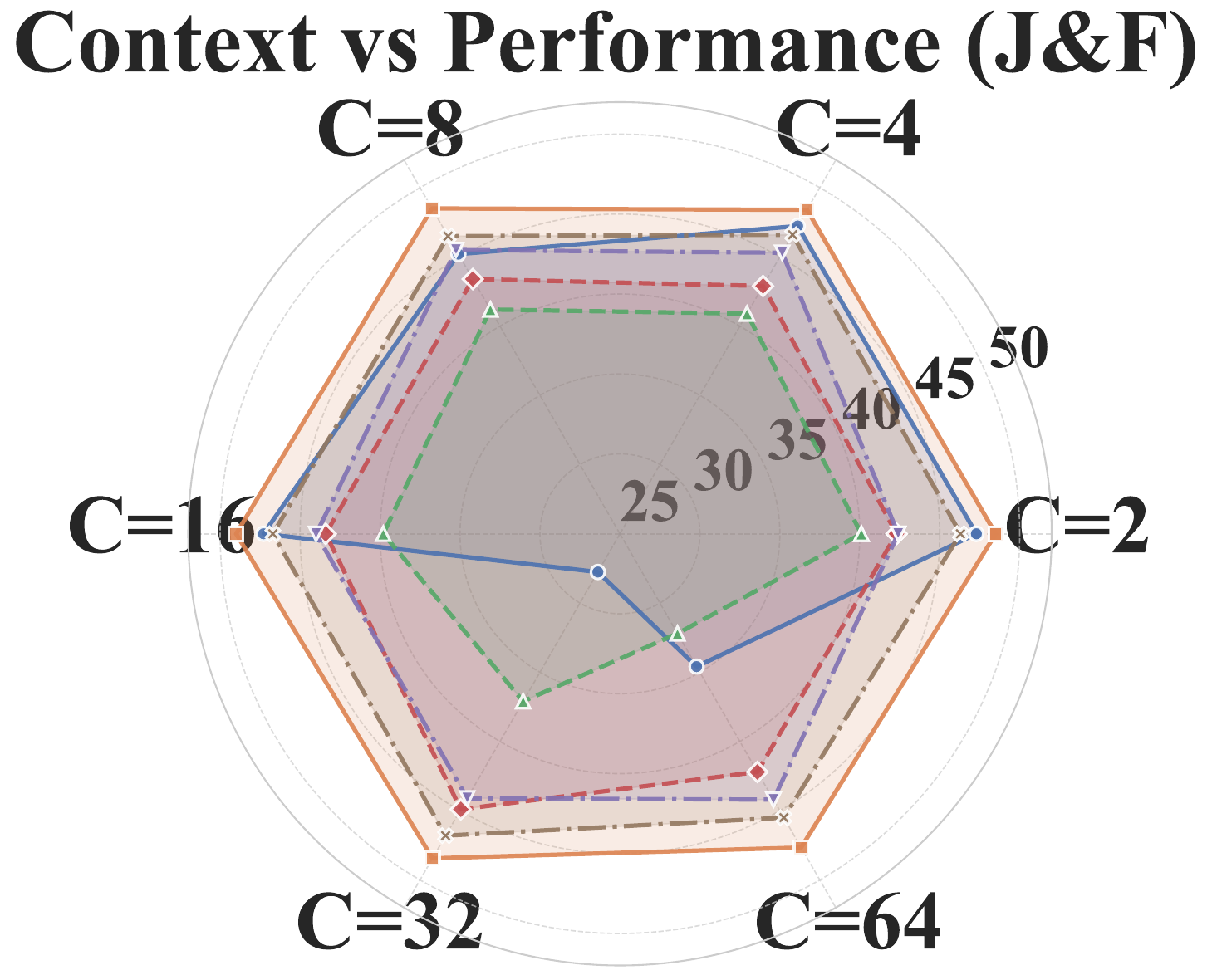}
        };
    \end{tikzpicture}
    \caption{Performance evaluation across compression methods. STAC maintains efficiency (main) and consistent performance across contexts (inset).}
    \label{fig:performance_eval}
    \vspace{-1.5em}
\end{figure}

\vspace{-1em}
\subsection{Ablation Studies}

All ablations follow the training recipe from \cref{ssec:exp_setup}, with 3K iteration training on referring data with context fixed at 32 frames. Validation is done on the held out ReasonVOS to analyze the zero-shot performance. 

\para{Compression strategy comparison.} \cref{fig:performance_eval} evaluates the compression approaches across context lengths from $C$=2 to 64 frames under standardized batch processing. The main plot illustrates the memory-latency trade-off, where VANILLA~\cite{lin2025glus} method exhibits quadratic scaling while compression methods maintain manageable footprints. The inset spider plot reveals performance that STAC consistently achieves the highest $\mathcal{J}$\&$\mathcal{F}$ scores with monotonic improvement as context increases, validating that our causal temporal design preserves long-range dependencies through selective state propagation. Perceiver~\cite{jaegle2021perceiver} saturates early due to its fixed query set, while Pooling plateaus from uniform averaging that discards motion information. BIMBA~\cite{islam2025bimba} remains competitive at short contexts but requires full video access, limiting streaming deployment. Crucially, in online scenarios, STAC maintains a constant inference latency of $\sim$200ms per frame regardless of sequence length, a capability inaccessible to bidirectional baselines that exhibit linear or quadratic growth.

\begin{table}[t]
\begin{minipage}[t]{0.62\linewidth}
\centering
\footnotesize
\captionof{table}{Evaluation of different directional combinations of spatial and temporal scanning with pruning and merging strategies.}
\label{tab:enrichment_timing}
\vspace{0.5ex}
\resizebox{\linewidth}{!}{%
\setlength{\tabcolsep}{3pt}%
\begin{tabular}{c|@{\hskip 8pt}c@{\hskip 12pt}c@{\hskip 8pt}|c|c|ccc}
\toprule
 \textbf{Spatial} & \multicolumn{2}{c|}{\textbf{Temporal Scan}} & \multirow{2}{*}{\textbf{Prune}} & \multirow{2}{*}{\textbf{Merge}} & \multicolumn{3}{c}{\textbf{ReasonVOS}} \\
 \cmidrule{2-3} \cmidrule{6-8}
 \textbf{Scan} & \textbf{Uni-Dir} & \textbf{Bi-Dir} & & & $\mathcal{J}$\&$\mathcal{F}$ & $\mathcal{J}$ & $\mathcal{F}$ \\
\midrule
\midrule
 \ding{55} & \ding{55} & \ding{55} & \ding{55} & \ding{55} & 48.1 & 45.4 & 50.0 \\
 \ding{51} & \ding{55} & \ding{55} & \ding{55} & \ding{51} & 44.1 & 41.0 & 47.1 \\
 \ding{51} & \ding{55} & \ding{51} & \ding{55} & \ding{51} & 42.7 & 39.4 & 46.1 \\
 \ding{55} & \ding{51} & \ding{55} & \ding{55} & \ding{51} & 45.2 & 42.2 & 48.3 \\
 \ding{51} & \ding{51} & \ding{55} & \ding{55} & \ding{51} & \textbf{49.6} & \textbf{47.2} & \textbf{52.0} \\
 \ding{51} & \ding{51} & \ding{55} & \ding{51} & \ding{55} & 47.9 & 45.4 & 50.4 \\
\bottomrule
\end{tabular}%
}
\end{minipage}%
\hfill
\begin{minipage}[t]{0.365\linewidth}
\centering
\footnotesize
\captionof{table}{Comparison of scanning and compression ordering (spatial vs.\ temporal-first).}
\label{tab:compression_order}
\vspace{0.5ex}
\resizebox{\linewidth}{!}{%
\begin{tabular}{cc|cc|ccc}
\toprule
 \multicolumn{2}{c|}{\textbf{Scan}} & \multicolumn{2}{c|}{\textbf{Comp.}} & \multicolumn{3}{c}{\textbf{ReasonVOS}} \\ \midrule
 \textbf{Sp.} & \textbf{Tmp.} & \textbf{Sp.} & \textbf{Tmp.} & \multirow{2}{*}{$\mathcal{J}$\&$\mathcal{F}$} & \multirow{2}{*}{$\mathcal{J}$} & \multirow{2}{*}{$\mathcal{F}$} \\
 \textbf{1st} & \textbf{1st} & \textbf{1st} & \textbf{1st} & & & \\
\midrule
           & \ding{51} & \ding{51} &           & 31.3 & 27.5 & 35.2 \\
           & \ding{51} &           & \ding{51} & 29.6 & 25.4 & 33.9 \\
 \ding{51} &           & \ding{51} &           & 45.7 & 43.4 & 48.0 \\
 \ding{51} &           &           & \ding{51} & \textbf{49.6} & \textbf{47.2} & \textbf{52.0} \\
\bottomrule
\end{tabular}%
}
\end{minipage}

\vspace{-3ex}
\end{table}

\para{Spatiotemporal enhancement architecture.} \cref{tab:enrichment_timing} validates our asymmetric scanning strategy, where bidirectional spatial scanning captures symmetric within-frame relationships while unidirectional temporal scanning enables streaming compatibility. Compared to without any enhancement or compression (48.1 $\mathcal{J}$\&$\mathcal{F}$), our full architecture reaches higher performance (49.6), confirming that globally-aware features  enable superior token selection. Bidirectional scanning on both dimensions yield lower performance while(42.7) at the same time eliminates streaming capacity and introduces future information leakage, while unidirectional on spatial and temporal scan (44.1) preserves causality but compromises spatial accuracy. Temporal-only unidirectional method (45.2) outperforms bidirectional variants yet underperforms our dual-stage design. Comparing aggregation strategies, merging (49.6) significantly outperforms pruning (47.9) by retaining information through averaging rather than elimination.

\para{Compression ordering.} \cref{tab:compression_order} evaluates the hierarchical sequence of our compression modules and reveals that temporal-first compression yields 49.6 $\mathcal{J}$\&$\mathcal{F}$ while significantly surpassing the 45.7$\mathcal{J}$\&$\mathcal{F}$achieved by spatial-first compression. This performance gap stems from the fact that spatial-first pooling prematurely eliminates the fine-grained inter-frame differences that constitute motion signals before the temporal scanner can identify salient regions. By contrast, our full four-stage pipeline consisting a cascade of spatial enrichment, temporal enrichment, temporal compression, and spatial compression optimizes the trade-off between context density and motion preservation. This specific sequence first captures motion patterns during the enrichment phase and prioritizes temporal reduction to ensure that critical dynamic features are preserved through the final spatial aggregation stage which architecturally enforces the principled separation of spatial context and causal temporal evolution.

\para{Online vs. offline inference.} With temporal recurrence being causal by design (\cref{eq:2},\cref{eq:3}), STAC runs identically in both modes, with only the EMA statistics (\cref{eq:adaptiveth}) warming up over the first few frames. The cost is small on referring queries, where the online STAC reaches 72.5~$\mathcal{J}\&\mathcal{F}$ on Ref-DAVIS17 against 73.6 offline. The gap widens on reasoning (48.8 vs.\ 52.3 on ReasonVOS), where queries rely on future frames and get updated as the frames come in.

\begin{figure*}[!t]
    \centering
    \includegraphics[width=1\linewidth]{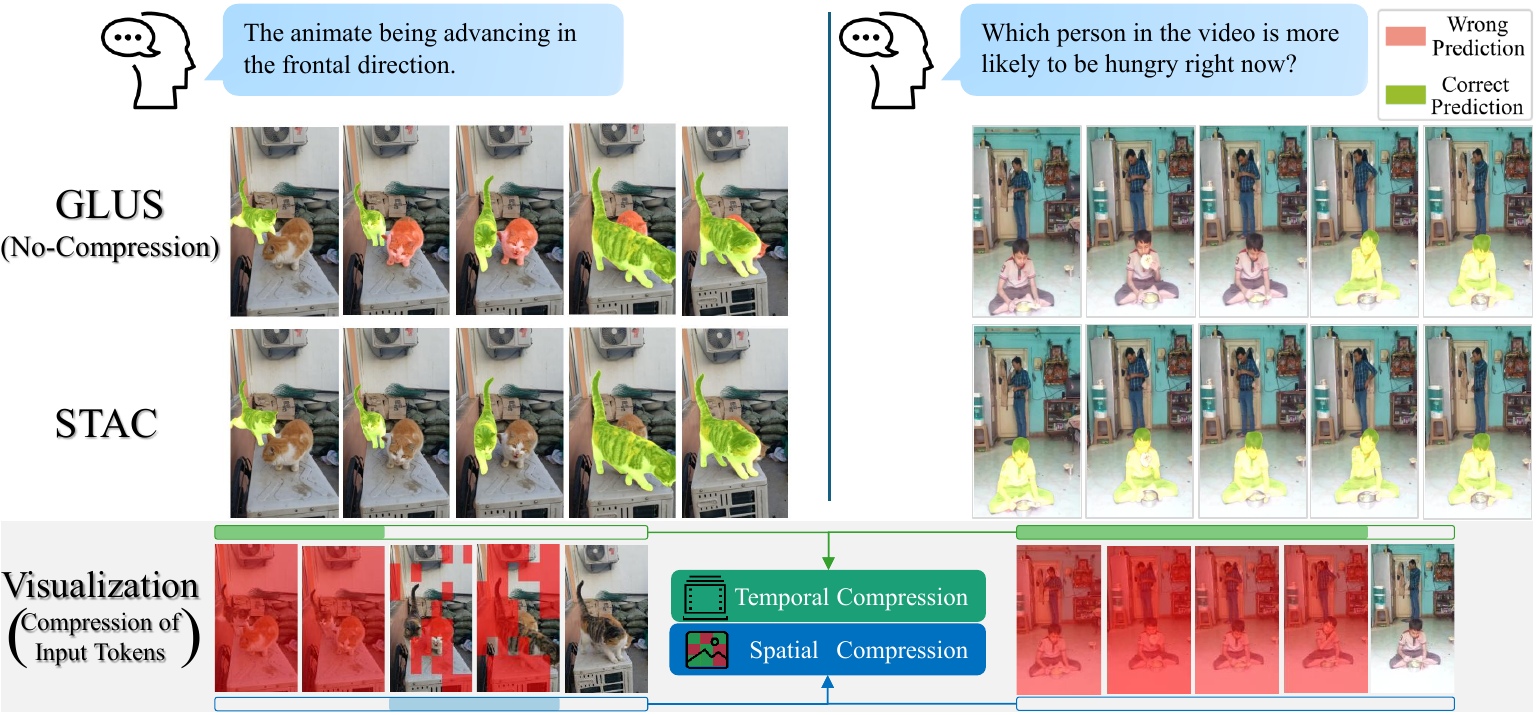}
    \caption{Performance on Segmentation with reasoning shows that our method performs aggressive yet adaptive compression. While the RED areas indicate compressed frames or patches, the use of selective aggregation propagates all the important information within those patches without missing any critial information.}
    \label{fig:viscomp}
    
  \vspace{-1.5ex}
\end{figure*}

\para{Qualitative analysis.} 
A qualitative comparison between GLUS and STAC on complex reasoning queries is shown in \cref{fig:viscomp}. Both methods successfully segment targets, but STAC maintains comparable quality while using only 15\% of tokens. For both queries, STAC correctly identifies and tracks the target throughout the sequence with greater temporal consistency than GLUS. The bottom visualization shows compression decisions, where red patches indicate discarded tokens. Temporal compression retains motion-rich frames while merging static ones, and spatial compression preserves foreground subjects while compressing the background. Through learned pooling, as discussed in \cref{sec:methodology}, thresholds adapt automatically to content characteristics, achieving over 90\% compression in static regions while retaining 30\%-50\% of tokens in dynamic sequences containing discriminative motion patterns essential for temporal reasoning.

\vspace{-0.5em}
\section{Conclusion}
\label{sec:conclusion}
\vspace{-0.4em}

This paper presents STAC, demonstrating that the redundancy signal intrinsic to SSM recurrence enables principled video token compression at linear cost. Exploiting this signal, STAC enriches features through selective state conditioning before compressing them, decoupling bidirectional spatial from causal temporal scanning to support streaming-compatible deployment, with task-grounded hierarchical compression to achieve 85\% token reduction and 1.8$\times$ speedup while surpassing compression-free baselines on reasoning segmentation benchmarks. Zero-shot transfer results further suggest that enrichment-informed compression learns generalisable representations beyond the training distribution. Future directions include quantifying the relationship between recurrent state capacity and achievable compression ratios, and extending the framework to multimodal streams such as audio-visual reasoning and embodied perception.

\section*{Acknowledgements}
This research work is supported by the Natural Science Foundation of China (No. 62576176), Agency for Science, Technology and Research (A*STAR) under its MTC Programmatic Funds (Grant No. M23L7b0021) and A*STAR Graduate Scholarship. The computational resources are supported by the Supercomputing Center of Nankai University (NKSC).

%
%
\bibliographystyle{splncs04}
\bibliography{main}

@String(CVPR= {IEEE Conf. Comput. Vis. Pattern Recog.})

@String(ICCV= {Int. Conf. Comput. Vis.})

@String(ECCV= {Eur. Conf. Comput. Vis.})

@String(NIPS= {Adv. Neural Inform. Process. Syst.})

@String(ICML = {Int. Conf. Machine Learning})

@String(ICLR = {Int. Conf. Learn. Represent.})

@String(AAAI = {AAAI})

@String(CVPR  = {CVPR})

@String(ICCV  = {ICCV})

@String(ECCV  = {ECCV})

@String(NIPS  = {NeurIPS})

@String(ICLR  = {ICLR})

@article{jiang2025token,
  title={Token-efficient long video understanding for multimodal llms},
  author={Jiang, Jindong and Li, Xiuyu and Liu, Zhijian and Li, Muyang and Chen, Guo and Li, Zhiqi and Huang, De-An and Liu, Guilin and Yu, Zhiding and Keutzer, Kurt and others},
  journal={arXiv preprint arXiv:2503.04130},
  year={2025}
}

@inproceedings{li2024llama,
  title={Llama-vid: An image is worth 2 tokens in large language models},
  author={Li, Yanwei and Wang, Chengyao and Jia, Jiaya},
  booktitle={European Conference on Computer Vision},
  pages={323--340},
  year={2024},
  organization={Springer}
}

@article{cheng2024videollama,
  title={Videollama 2: Advancing spatial-temporal modeling and audio understanding in video-llms},
  author={Cheng, Zesen and Leng, Sicong and Zhang, Hang and Xin, Yifei and Li, Xin and Chen, Guanzheng and Zhu, Yongxin and Zhang, Wenqi and Luo, Ziyang and Zhao, Deli and others},
  journal={arXiv preprint arXiv:2406.07476},
  year={2024}
}

@article{touvron2023llama,
  title={Llama: Open and efficient foundation language models},
  author={Touvron, Hugo and Lavril, Thibaut and Izacard, Gautier and Martinet, Xavier and Lachaux, Marie-Anne and Lacroix, Timoth{\'e}e and Rozi{\`e}re, Baptiste and Goyal, Naman and Hambro, Eric and Azhar, Faisal and others},
  journal={arXiv preprint arXiv:2302.13971},
  year={2023}
}

@inproceedings{hu2025m,
  title={M-LLM based video frame selection for efficient video understanding},
  author={Hu, Kai and Gao, Feng and Nie, Xiaohan and Zhou, Peng and Tran, Son and Neiman, Tal and Wang, Lingyun and Shah, Mubarak and Hamid, Raffay and Yin, Bing and others},
  booktitle=CVPR,
  pages={13702--13712},
  year={2025}
}

@article{tang2025tspo,
  title={TSPO: Temporal Sampling Policy Optimization for Long-form Video Language Understanding},
  author={Tang, Canhui and Han, Zifan and Sun, Hongbo and Zhou, Sanping and Zhang, Xuchong and Wei, Xin and Yuan, Ye and Xu, Jinglin and Sun, Hao},
  journal={arXiv preprint arXiv:2508.04369},
  year={2025}
}

@inproceedings{zhu2024vision,
  title={Vision mamba: Efficient visual representation learning with bidirectional state space model},
  author={Zhu, Lianghui and Liao, Bencheng and Zhang, Qian and Wang, Xinlong and Liu, Wenyu and Wang, Xinggang},
  booktitle=ICML,
  pages={62429--62442},
  year={2024},
}

@inproceedings{li2024videomamba,
  title={Videomamba: State space model for efficient video understanding},
  author={Li, Kunchang and Li, Xinhao and Wang, Yi and He, Yinan and Wang, Yali and Wang, Limin and Qiao, Yu},
  booktitle=ECCV,
  pages={237--255},
  year={2024},
  organization={Springer}
}

@inproceedings{park2024videomamba,
  title={Videomamba: Spatio-temporal selective state space model},
  author={Park, Jinyoung and Kim, Hee-Seon and Ko, Kangwook and Kim, Minbeom and Kim, Changick},
  booktitle=ECCV,
  pages={1--18},
  year={2024},
  organization={Springer}
}

@article{yang2024vivim,
  title={Vivim: A video vision mamba for medical video segmentation},
  author={Yang, Yijun and Xing, Zhaohu and Yu, Lequan and Huang, Chunwang and Fu, Huazhu and Zhu, Lei},
  journal={arXiv preprint arXiv:2401.14168},
  year={2024}
}

@article{chen2024video,
  title={Video mamba suite: State space model as a versatile alternative for video understanding},
  author={Chen, Guo and Huang, Yifei and Xu, Jilan and Pei, Baoqi and Chen, Zhe and Li, Zhiqi and Wang, Jiahao and Li, Kunchang and Lu, Tong and Wang, Limin},
  journal={arXiv preprint arXiv:2403.09626},
  year={2024}
}

@inproceedings{pei2025efficientvmamba,
  title={Efficientvmamba: Atrous selective scan for light weight visual mamba},
  author={Pei, Xiaohuan and Huang, Tao and Xu, Chang},
  booktitle=AAAI,
  pages={6443--6451},
  year={2025}
}

@inproceedings{luo2023soc,
  title={Soc: Semantic-assisted object cluster for referring video object segmentation},
  author={Luo, Zhuoyan and Xiao, Yicheng and Liu, Yong and Li, Shuyan and Wang, Yitong and Tang, Yansong and Li, Xiu and Yang, Yujiu},
  booktitle=NIPS,
  volume={36},
  pages={26425--26437},
  year={2023}
}

@inproceedings{lin2025glus,
  title={Glus: Global-local reasoning unified into a single large language model for video segmentation},
  author={Lin, Lang and Yu, Xueyang and Pang, Ziqi and Wang, Yu-Xiong},
  booktitle=CVPR,
  pages={8658--8667},
  year={2025}
}

@inproceedings{bai2024one,
  title={One token to seg them all: Language instructed reasoning segmentation in videos},
  author={Bai, Zechen and He, Tong and Mei, Haiyang and Wang, Pichao and Gao, Ziteng and Chen, Joya and Zhang, Zheng and Shou, Mike Zheng},
  booktitle=NIPS,
  pages={6833--6859},
  year={2024}
}

@inproceedings{kirillov2023segany,
  title={Segment anything},
  author={Kirillov, Alexander and Mintun, Eric and Ravi, Nikhila and Mao, Hanzi and Rolland, Chloe and Gustafson, Laura and Xiao, Tete and Whitehead, Spencer and Berg, Alexander C and Lo, Wan-Yen and others},
  booktitle=ICCV,
  pages={4015--4026},
  year={2023}
}

@inproceedings{miao2023spectrum,
  title={Spectrum-guided multi-granularity referring video object segmentation},
  author={Miao, Bo and Bennamoun, Mohammed and Gao, Yongsheng and Mian, Ajmal},
  booktitle={Proceedings of the IEEE/CVF International Conference on Computer Vision},
  pages={920--930},
  year={2023}
}

@article{ravi2024sam,
  title={Sam 2: Segment anything in images and videos},
  author={Ravi, Nikhila and Gabeur, Valentin and Hu, Yuan-Ting and Hu, Ronghang and Ryali, Chaitanya and Ma, Tengyu and Khedr, Haitham and R{\"a}dle, Roman and Rolland, Chloe and Gustafson, Laura and others},
  journal={arXiv preprint arXiv:2408.00714},
  year={2024}
}

@inproceedings{rasheed2024glamm,
  title={Glamm: Pixel grounding large multimodal model},
  author={Rasheed, Hanoona and Maaz, Muhammad and Shaji, Sahal and Shaker, Abdelrahman and Khan, Salman and Cholakkal, Hisham and Anwer, Rao M and Xing, Eric and Yang, Ming-Hsuan and Khan, Fahad S},
  booktitle={Proceedings of the IEEE/CVF Conference on Computer Vision and Pattern Recognition},
  pages={13009--13018},
  year={2024}
}

@inproceedings{lai2024lisa,
  title={Lisa: Reasoning segmentation via large language model},
  author={Lai, Xin and Tian, Zhuotao and Chen, Yukang and Li, Yanwei and Yuan, Yuhui and Liu, Shu and Jia, Jiaya},
  booktitle=CVPR,
  pages={9579--9589},
  year={2024}
}

@inproceedings{radford2021learning,
  title={Learning transferable visual models from natural language supervision},
  author={Radford, Alec and Kim, Jong Wook and Hallacy, Chris and Ramesh, Aditya and Goh, Gabriel and Agarwal, Sandhini and Sastry, Girish and Askell, Amanda and Mishkin, Pamela and Clark, Jack and others},
  booktitle=ICML,
  pages={8748--8763},
  year={2021},
  organization={PmLR}
}

@misc{liu2024llavanext,
  title={{LLaVA-NeXT}: Improved reasoning, {OCR}, and world knowledge},
  url={https://llava-vl.github.io/blog/2024-01-30-llava-next/},
  author={Liu, Haotian and Li, Chunyuan and Li, Yuheng and Li, Bo and Zhang, Yuanhan and Shen, Sheng and Lee, Yong Jae},
  month={January},
  year={2024},
  note={(Accessed: 2026-06-25)}
}

@inproceedings{liu2023visual,
  title={Visual instruction tuning},
  author={Liu, Haotian and Li, Chunyuan and Wu, Qingyang and Lee, Yong Jae},
  booktitle=NIPS,
  volume={36},
  pages={34892--34916},
  year={2023}
}

@article{bai2023qwenvl,
  title={Qwen-vl: A frontier large vision-language model with versatile abilities},
  author={Bai, Jinze and Bai, Shuai and Yang, Shusheng and Wang, Shijie and Tan, Sinan and Wang, Peng and Lin, Junyang and Zhou, Chang and Zhou, Jingren},
  journal={arXiv preprint arXiv:2308.12966},
  year={2023}
}

@inproceedings{zhai2023sigmoid,
  title={Sigmoid loss for language image pre-training},
  author={Zhai, Xiaohua and Mustafa, Basil and Kolesnikov, Alexander and Beyer, Lucas},
  booktitle=ICCV,
  pages={11975--11986},
  year={2023}
}

@inproceedings{song2024moviechat,
  title={Moviechat: From dense token to sparse memory for long video understanding},
  author={Song, Enxin and Chai, Wenhao and Wang, Guanhong and Zhang, Yucheng and Zhou, Haoyang and Wu, Feiyang and Chi, Haozhe and Guo, Xun and Ye, Tian and Zhang, Yanting and others},
  booktitle=CVPR,
  pages={18221--18232},
  year={2024}
}

@inproceedings{jaegle2021perceiver,
  title={Perceiver: General perception with iterative attention},
  author={Jaegle, Andrew and Gimeno, Felix and Brock, Andy and Vinyals, Oriol and Zisserman, Andrew and Carreira, Joao},
  booktitle=ICML,
  pages={4651--4664},
  year={2021},
  organization={PMLR}
}

@article{bolya2022token,
  title={Token merging: Your vit but faster},
  author={Bolya, Daniel and Fu, Cheng-Yang and Dai, Xiaoliang and Zhang, Peizhao and Feichtenhofer, Christoph and Hoffman, Judy},
  journal={arXiv preprint arXiv:2210.09461},
  year={2022}
}

@inproceedings{rao2021dynamicvit,
  title={Dynamicvit: Efficient vision transformers with dynamic token sparsification},
  author={Rao, Yongming and Zhao, Wenliang and Liu, Benlin and Lu, Jiwen and Zhou, Jie and Hsieh, Cho-Jui},
  booktitle=NIPS,
  volume={34},
  pages={13937--13949},
  year={2021}
}

@article{liang2022not,
  title={Not all patches are what you need: Expediting vision transformers via token reorganizations},
  author={Liang, Youwei and Ge, Chongjian and Tong, Zhan and Song, Yibing and Wang, Jue and Xie, Pengtao},
  journal={arXiv preprint arXiv:2202.07800},
  year={2022}
}

@inproceedings{chen2024image,
  title={An image is worth 1/2 tokens after layer 2: Plug-and-play inference acceleration for large vision-language models},
  author={Chen, Liang and Zhao, Haozhe and Liu, Tianyu and Bai, Shuai and Lin, Junyang and Zhou, Chang and Chang, Baobao},
  booktitle=ECCV,
  pages={19--35},
  year={2024},
  organization={Springer}
}

@inproceedings{ye2024atpllava,
  title={Atp-llava: Adaptive token pruning for large vision language models},
  author={Ye, Xubing and Gan, Yukang and Ge, Yixiao and Zhang, Xiao-Ping and Tang, Yansong},
  booktitle=CVPR,
  pages={24972--24982},
  year={2025}
}

@article{shen2024tempme,
  title={Tempme: Video temporal token merging for efficient text-video retrieval},
  author={Shen, Leqi and Hao, Tianxiang and He, Tao and Zhao, Sicheng and Zhang, Yifeng and Liu, Pengzhang and Bao, Yongjun and Ding, Guiguang},
  journal={arXiv preprint arXiv:2409.01156},
  year={2024}
}

@article{shen2024longvu,
  title={Longvu: Spatiotemporal adaptive compression for long video-language understanding},
  author={Shen, Xiaoqian and Xiong, Yunyang and Zhao, Changsheng and Wu, Lemeng and Chen, Jun and Zhu, Chenchen and Liu, Zechun and Xiao, Fanyi and Varadarajan, Balakrishnan and Bordes, Florian and others},
  journal={arXiv preprint arXiv:2410.17434},
  year={2024}
}

@inproceedings{korbar2019scsampler,
  title={Scsampler: Sampling salient clips from video for efficient action recognition},
  author={Korbar, Bruno and Tran, Du and Torresani, Lorenzo},
  booktitle=ICCV,
  pages={6232--6242},
  year={2019}
}

@article{zhang2023video,
  title={Video-llama: An instruction-tuned audio-visual language model for video understanding},
  author={Zhang, Hang and Li, Xin and Bing, Lidong},
  journal={arXiv preprint arXiv:2306.02858},
  year={2023}
}

@article{team2023gemini,
  title={Gemini: a family of highly capable multimodal models},
  author={Team, Gemini and Anil, Rohan and Borgeaud, Sebastian and Alayrac, Jean-Baptiste and Yu, Jiahui and Soricut, Radu and Schalkwyk, Johan and Dai, Andrew M and Hauth, Anja and Millican, Katie and others},
  journal={arXiv preprint arXiv:2312.11805},
  year={2023}
}

@inproceedings{wang2024internvideo2,
  title={Internvideo2: Scaling foundation models for multimodal video understanding},
  author={Wang, Yi and Li, Kunchang and Li, Xinhao and Yu, Jiashuo and He, Yinan and Chen, Guo and Pei, Baoqi and Zheng, Rongkun and Wang, Zun and Shi, Yansong and others},
  booktitle={European Conference on Computer Vision},
  pages={396--416},
  year={2024},
  organization={Springer}
}

@article{lin2023videollava,
  title={Video-LLaVA: Learning United Visual Representation by Alignment Before Projection},
  author={Lin, Bin and Zhu, Bin and Ye, Yang and Ning, Munan and Jin, Peng and Yuan, Li},
  journal={arXiv preprint arXiv:2311.10122},
  year={2023}
}

@inproceedings{gu2024mamba,
  title={Mamba: Linear-time sequence modeling with selective state spaces},
  author={Gu, Albert and Dao, Tri},
  booktitle={First Conference on Language Modeling},
  year={2024}
}

@article{gu2021efficiently,
  title={Efficiently modeling long sequences with structured state spaces},
  author={Gu, Albert and Goel, Karan and R{\'e}, Christopher},
  journal={arXiv preprint arXiv:2111.00396},
  year={2021}
}

@article{liu2024vmamba,
  title={Vmamba: Visual state space model},
  author={Liu, Yue and Tian, Yunjie and Zhao, Yuzhong and Yu, Hongtian and Xie, Lingxi and Wang, Yaowei and Ye, Qixiang and Jiao, Jianbin and Liu, Yunfan},
  journal=NIPS,
  volume={37},
  pages={103031--103063},
  year={2024}
}

@article{dao2024transformers,
  title={Transformers are ssms: Generalized models and efficient algorithms through structured state space duality},
  author={Dao, Tri and Gu, Albert},
  journal={arXiv preprint arXiv:2405.21060},
  year={2024}
}

@article{lu2024videomambapro,
  title={Videomambapro: A leap forward for mamba in video understanding},
  author={Lu, Hui and Salah, Albert Ali and Poppe, Ronald},
  journal={arXiv e-prints},
  pages={arXiv--2406},
  year={2024}
}

@inproceedings{yan2024visa,
  title={Visa: Reasoning video object segmentation via large language models},
  author={Yan, Cilin and Wang, Haochen and Yan, Shilin and Jiang, Xiaolong and Hu, Yao and Kang, Guoliang and Xie, Weidi and Gavves, Efstratios},
  booktitle=ECCV,
  pages={98--115},
  year={2024},
  organization={Springer}
}

@inproceedings{he2024decoupling,
  title={Decoupling static and hierarchical motion perception for referring video segmentation},
  author={He, Shuting and Ding, Henghui},
  booktitle=CVPR,
  pages={13332--13341},
  year={2024}
}

@inproceedings{wu2023onlinerefer,
  title={Onlinerefer: A simple online baseline for referring video object segmentation},
  author={Wu, Dongming and Wang, Tiancai and Zhang, Yuang and Zhang, Xiangyu and Shen, Jianbing},
  booktitle=ICCV,
  pages={2761--2770},
  year={2023}
}

@inproceedings{wu2022language,
  title={Language as queries for referring video object segmentation},
  author={Wu, Jiannan and Jiang, Yi and Sun, Peize and Yuan, Zehuan and Luo, Ping},
  booktitle=CVPR,
  pages={4974--4984},
  year={2022}
}

@inproceedings{botach2022endtoend,
  title={End-to-end referring video object segmentation with multimodal transformers},
  author={Botach, Adam and Zheltonozhskii, Evgenii and Baskin, Chaim},
  booktitle=CVPR,
  pages={4985--4995},
  year={2022}
}

@article{xu2018youtubevos,
  title={Youtube-vos: A large-scale video object segmentation benchmark},
  author={Xu, Ning and Yang, Linjie and Fan, Yuchen and Yue, Dingcheng and Liang, Yuchen and Yang, Jianchao and Huang, Thomas},
  journal={arXiv preprint arXiv:1809.03327},
  year={2018}
}

@inproceedings{ding2023mevis,
  title={MeViS: A large-scale benchmark for video segmentation with motion expressions},
  author={Ding, Henghui and Liu, Chang and He, Shuting and Jiang, Xudong and Loy, Chen Change},
  booktitle=ICCV,
  pages={2694--2703},
  year={2023}
}

@article{pont20172017,
  title={The 2017 davis challenge on video object segmentation},
  author={Pont-Tuset, Jordi and Perazzi, Federico and Caelles, Sergi and Arbel{\'a}ez, Pablo and Sorkine-Hornung, Alex and Van Gool, Luc},
  journal={arXiv preprint arXiv:1704.00675},
  year={2017}
}

@inproceedings{munasinghe2025videoglamm,
  title={Videoglamm: A large multimodal model for pixel-level visual grounding in videos},
  author={Munasinghe, Shehan and Gani, Hanan and Zhu, Wenqi and Cao, Jiale and Xing, Eric and Khan, Fahad Shahbaz and Khan, Salman},
  booktitle=CVPR,
  pages={19036--19046},
  year={2025}
}

@article{zhu2023tracking,
  title={Tracking with human-intent reasoning},
  author={Zhu, Jiawen and Cheng, Zhi-Qi and He, Jun-Yan and Li, Chenyang and Luo, Bin and Lu, Huchuan and Geng, Yifeng and Xie, Xuansong},
  journal={arXiv preprint arXiv:2312.17448},
  year={2023}
}

@inproceedings{gavrilyuk2018actor,
  title={Actor and action video segmentation from a sentence},
  author={Gavrilyuk, Kirill and Ghodrati, Amir and Li, Zhenyang and Snoek, Cees GM},
  booktitle=CVPR,
  pages={5958--5966},
  year={2018}
}

@article{yuan2025sa2va,
  title={Sa2va: Marrying sam2 with llava for dense grounded understanding of images and videos},
  author={Yuan, Haobo and Li, Xiangtai and Zhang, Tao and Huang, Zilong and Xu, Shilin and Ji, Shunping and Tong, Yunhai and Qi, Lu and Feng, Jiashi and Yang, Ming-Hsuan},
  journal={arXiv preprint arXiv:2501.04001},
  year={2025}
}

@inproceedings{khoreva2018video,
  title={Video object segmentation with language referring expressions},
  author={Khoreva, Anna and Rohrbach, Anna and Schiele, Bernt},
  booktitle={Asian conference on computer vision},
  pages={123--141},
  year={2018},
  organization={Springer}
}

@inproceedings{tang2023temporal,
  title={Temporal collection and distribution for referring video object segmentation},
  author={Tang, Jiajin and Zheng, Ge and Yang, Sibei},
  booktitle=ICCV,
  pages={15466--15476},
  year={2023}
}

@inproceedings{seo2020urvos,
  title={Urvos: Unified referring video object segmentation network with a large-scale benchmark},
  author={Seo, Seonguk and Lee, Joon-Young and Han, Bohyung},
  booktitle=ECCV,
  pages={208--223},
  year={2020},
  organization={Springer}
}

@inproceedings{ding2022language,
  title={Language-bridged spatial-temporal interaction for referring video object segmentation},
  author={Ding, Zihan and Hui, Tianrui and Huang, Junshi and Wei, Xiaoming and Han, Jizhong and Liu, Si},
  booktitle=CVPR,
  pages={4964--4973},
  year={2022}
}

@inproceedings{islam2025bimba,
  title={Bimba: Selective-scan compression for long-range video question answering},
  author={Islam, Md Mohaiminul and Nagarajan, Tushar and Wang, Huiyu and Bertasius, Gedas and Torresani, Lorenzo},
  booktitle=CVPR,
  pages={29096--29107},
  year={2025}
}

@article{chen2024longvila,
  title={Longvila: Scaling long-context visual language models for long videos},
  author={Chen, Yukang and Xue, Fuzhao and Li, Dacheng and Hu, Qinghao and Zhu, Ligeng and Li, Xiuyu and Fang, Yunhao and Tang, Haotian and Yang, Shang and Liu, Zhijian and others},
  journal={arXiv preprint arXiv:2408.10188},
  year={2024}
}

@article{bengio2013estimating,
  title={Estimating or propagating gradients through stochastic neurons for conditional computation},
  author={Bengio, Yoshua and L{\'e}onard, Nicholas and Courville, Aaron},
  journal={arXiv preprint arXiv:1308.3432},
  year={2013}
}

@article{dubey2024llama,
  title={The llama 3 herd of models},
  author={Dubey, Abhimanyu and Jauhri, Abhinav and Pandey, Abhinav and Kadian, Abhishek and Al-Dahle, Ahmad and Letman, Aiesha and Mathur, Akhil and Schelten, Alan and Yang, Amy and Fan, Angela and others},
  journal={arXiv e-prints},
  pages={arXiv--2407},
  year={2024}
}

@inproceedings{hu2022lora,
  title={{LoRA}: Low-Rank Adaptation of Large Language Models},
  author={Edward J Hu and Yelong Shen and Phillip Wallis and Zeyuan Allen-Zhu and Yuanzhi Li and Shean Wang and Lu Wang and Weizhu Chen},
  booktitle=ICLR,
  year={2022}
}

@article{achiam2023gpt,
  title={Gpt-4 technical report},
  author={Achiam, Josh and Adler, Steven and Agarwal, Sandhini and Ahmad, Lama and Akkaya, Ilge and Aleman, Florencia Leoni and Almeida, Diogo and Altenschmidt, Janko and Altman, Sam and Anadkat, Shyamal and others},
  journal={arXiv preprint arXiv:2303.08774},
  year={2023}
}

@article{touvron2023llama2,
    title={Llama 2: Open foundation and fine-tuned chat models},
    author={Touvron, Hugo and Martin, Louis and Stone, Kevin and Albert, Peter and Almahairi, Amjad and Babaei, Yasmine and Bashlykov, Nikolay and Batra, Soumya and Bhargava, Prajjwal and Bhosale, Shruti and others},
    journal={arXiv:2307.09288},
    year={2023}
}

@article{jiang2024mixtral,
    title={Mixtral of experts},
    author={Jiang, Albert Q and Sablayrolles, Alexandre and Roux, Antoine and Mensch, Arthur and Savary, Blanche and Bamford, Chris and Chaplot, Devendra Singh and Casas, Diego de las and Hanna, Emma Bou and Bressand, Florian and others},
    journal={arXiv:2401.04088},
    year={2024}
}

@article{li2024llava,
  title={LLaVA-OneVision: Easy Visual Task Transfer},
  author={Li, Bo and Zhang, Yuanhan and Guo, Dong and Zhang, Renrui and Li, Feng and Zhang, Hao and Zhang, Kaichen and Li, Yanwei and Liu, Ziwei and Li, Chunyuan},
  journal={arXiv preprint arXiv:2408.03326},
  year={2024}
}

@article{chen2023internvl,
  title={Internvl: Scaling up vision foundation models and aligning for generic visual-linguistic tasks},
  author={Chen, Zhe and Wu, Jiannan and Wang, Wenhai and Su, Weijie and Chen, Guo and Xing, Sen and Muyan, Zhong and Zhang, Qinglong and Zhu, Xizhou and Lu, Lewei and others},
  journal={arXiv preprint arXiv:2312.14238},
  year={2023}
}

@inproceedings{gu2020hippo,
  title={HiPPO: Recurrent Memory with Optimal Polynomial Projections},
  author={Gu, Albert and Dao, Tri and Ermon, Stefano and Rudra, Atri and R{\'e}, Christopher},
  booktitle=NIPS,
  volume={33},
  year={2020}
}

@inproceedings{gu2021combining,
  title={Combining Recurrent, Convolutional, and Continuous-time Models with Linear State-Space Layers},
  author={Gu, Albert and Johnson, Isys and Goel, Karan and Saab, Khaled and Dao, Tri and Rudra, Atri and R{\'e}, Christopher},
  booktitle=NIPS,
  volume={34},
  year={2021}
}

@inproceedings{gu2022efficiently,
  title={Efficiently Modeling Long Sequences with Structured State Spaces},
  author={Gu, Albert and Goel, Karan and R\'e, Christopher},
  booktitle=ICLR,
  year={2022}
}

@inproceedings{gu2022s4d,
  title={On the Parameterization and Initialization of Diagonal State Space Models},
  author={Gu, Albert and Gupta, Ankit and Goel, Karan and R\'e, Christopher},
  booktitle=NIPS,
  volume={35},
  year={2022}
}

@inproceedings{pan2021ia,
  title={IA-RED 2: Interpretability-aware redundancy reduction for vision transformers},
  author={Pan, Bowen and Panda, Rameswar and Jiang, Yifan and Wang, Zhangyang and Feris, Rogerio and Oliva, Aude},
  booktitle=NIPS,
  volume={34},
  pages={24898--24911},
  year={2021}
}

@inproceedings{zhao2025cobra,
  title={Cobra: Extending mamba to multi-modal large language model for efficient inference},
  author={Zhao, Han and Zhang, Min and Zhao, Wei and Ding, Pengxiang and Huang, Siteng and Wang, Donglin},
  booktitle=AAAI,
  pages={10421--10429},
  year={2025}
}

@article{kalman1960new,
  title={A new approach to linear filtering and prediction problems},
  author={Kalman, Rudolph Emil},
  journal={Journal of Basic Engineering},
  volume={82},
  number={1},
  pages={35--45},
  year={1960}
}

@article{adam2014method,
  title={A method for stochastic optimization},
  author={Adam, Kingma DP Ba J and others},
  journal={arXiv preprint arXiv:1412.6980},
  volume={1412},
  number={6},
  year={2014}
}

@inproceedings{gong2025devil,
  title={The devil is in temporal token: High quality video reasoning segmentation},
  author={Gong, Sitong and Zhuge, Yunzhi and Zhang, Lu and Yang, Zongxin and Zhang, Pingping and Lu, Huchuan},
  booktitle={Proceedings of the IEEE/CVF Conference on Computer Vision and Pattern Recognition},
  pages={29183--29192},
  year={2025}
}

@article{liu2024vision,
  title={Vision transformers with hierarchical attention},
  author={Liu, Yun and Wu, Yu-Huan and Sun, Guolei and Zhang, Le and Chhatkuli, Ajad and Van Gool, Luc},
  journal={Machine Intelligence Research},
  volume={21},
  number={4},
  pages={670--683},
  year={2024},
  publisher={Springer}
}

@article{ariff2026evaluating,
  title={Evaluating SAM2 for Video Semantic Segmentation},
  author={Ariff, Syed Hesham Syed and Liu, Yun and Sun, Guolei and Yang, Jing and Ding, Henghui and Geng, Xue and Jiang, Xudong},
  journal={Machine Intelligence Research},
  year={2026}
}

@article{ning2026video,
  title={Video-Bench: A Comprehensive Benchmark and Toolkit for Evaluating Video-Based Large Language Models}, 
  author={Ning, Munan and Zhu, Bin and Xie, Yujia and Lin, Bin and Cui, Jiaxi and Yuan, Lu and Chen, Dongdong and Yuan, Li},
  journal={Computational Visual Media}, 
  volume={12},
  number={1},
  pages={71--84},
  year={2026},
  publisher={Tsinghua University Press}
}

@article{karacan2025full,
  title={Full-Frame Video Stabilization via Spatiotemporal Transformers}, 
  author={Karacan, Levent and Sarıgül, Mehmet},
  journal={Computational Visual Media}, 
  volume={11},
  number={3},
  pages={655--667},
  year={2025},
  publisher={Tsinghua University Press}
}

@article{han2026object,
  title = {Object-Centric Video Prediction with Mask-Guided Spatiotemporal Diffusion},
  author = {Chenchen Han and Jiayi Fan and Na Wu and Jinpeng Dai and Hanbin Bao and Xiankai Lu},
  journal = {Machine Intelligence Research},
  year = {2026},
  publisher={Springer}
}

@article{an2026video,
  title = {Video Understanding: From Geometry and Semantics to Unified Models},
  author = {Zhaochong An and Zirui Li and Mingqiao Ye and Feng Qiao and Jiaang Li and Zongwei Wu and Vishal Thengane and Chengzu Li and Lei Li and Luc Van Gool and Guolei Sun and Serge Belongie},
  journal = {Machine Intelligence Research},
  year = {2026},
  publisher={Springer}
}

@article{wang2026SAMDistill,
  title = {{SAMDistill}: {SAM}-based Spatial-temporal Distillation for Robust 3D Object Detection},
  author = {Zhaozhong Wang and Dian Shao and Lei Zhang and Zuowei Zhang and Binglu Wang},
  journal = {Machine Intelligence Research},
  year = {2026},
  publisher={Springer}
}

@article{feng2026training,
  title = {Training-free Dense Video Captioning with Large-scale Pre-trained Models},
  author = {Yue Feng and Ziyi Yan and Yizhen Jia and Ethan Q. Chen and Jie Qin},
  journal = {Machine Intelligence Research},
  year = {2026},
  publisher={Springer}
}

\end{document}


\title{\texorpdfstring{\scalebox{1.8}{\twemoji{books}}}{} STAC: Selective Spatiotemporal Aggregation and Compression for Video Reasoning Segmentation}

\titlerunning{STAC}


\author{%
Syed Ariff Syed Hesham\textsuperscript{1,2}, 
Yun Liu\textsuperscript{3,4,5}\thanks{Corresponding author: Yun Liu (liuyun@nankai.edu.cn)}, 
Guolei Sun\textsuperscript{3,4,5}, 
Jing Yang\textsuperscript{6}, \\ 
Henghui Ding\textsuperscript{7}, 
Xue Geng\textsuperscript{2}, 
Xudong Jiang\textsuperscript{1}  \\
}

\authorrunning{~Hesham et al.}

\institute{School of EEE, Nanyang Technological University, Singapore \and
Institute for Infocomm Research, A*STAR, Singapore
 \and
VCIP, CS, Nankai University, Tianjin, China \and
AAIS, Nankai University, Tianjin, China \and
NKIARI, Shenzhen Futian, China \and
Guizhou University, Guizhou, China \and
Fudan University, Shanghai, China
}

\clearpage
\setcounter{page}{1}
\maketitlesupplementary

\setcounter{section}{0}
\setcounter{figure}{0}
\setcounter{table}{0}
\setcounter{equation}{0}

\renewcommand{\thesection}{A.\arabic{section}}
\renewcommand{\thefigure}{A.\arabic{figure}}
\renewcommand{\thetable}{A.\arabic{table}}
\renewcommand{\theequation}{A.\arabic{equation}}

\renewcommand{\thesubsection}{A.\arabic{section}.\arabic{subsection}}

\section{Additional Implementation Details}
\label{sec:supp_implementation}
\subsection{Architecture and Training Configuration}

\para{Vision Encoder and Spatiotemporal Aggregation.} Video frames are individually processed through a pre-trained CLIP ViT-L/14~\cite{radford2021learning} to get the frame-level encoded features $\mathbf{F}_e \in \mathbb{R}^{T \times N \times d_e}$. The encoded features are then passed through our novel State-informed Spatiotemporal Aggregator (SSA) that enriches the encoder features with spatiotemporal context. The SSA module then processes these features through two sequential stages, starting with a spatial aggregation stage that applies bidirectional state scanning~\cite{zhu2024vision} to capture symmetric within-frame relationships, processing all $T$ frames independently in parallel, then through the causal spatially parallelized scanning that accumulates the temporal context. This process could be represented as,

\begin{equation}
\mathbf{F}_{\text{spatial}}[t] = \frac{1}{2}\left(\mathcal{M}^b_{\rightarrow}(\mathbf{F}_e[t]) + \mathcal{M}^b_{\leftarrow}(\mathbf{F}_e[t])\right),
\end{equation}

\noindent where the causal selective state-space mechanism accumulates \cite{dao2024transformers} temporal context through the operation, 
\begin{equation}
\label{eq:recurrentST}
\begin{aligned}
\mathbf{h}_{n,t} &= \bar{\mathbf{A}}_t \mathbf{h}_{n,t-1} + \bar{\mathbf{B}}_t \mathbf{F}_{\text{spatial}}[t,n], \\
\mathbf{F}_{\text{ST}}[t,n] &= \bar{\mathbf{C}}_t \mathbf{h}_{n,t}, \quad n = 1, \ldots, N.
\end{aligned}
\end{equation}

\noindent Here, $\mathcal{M}^b_{\rightarrow}$ and $\mathcal{M}^b_{\leftarrow}$ are the forward and backward scanning in the bidirectional selective scan operation, and $\mathcal{M}_{\text{causal}}$ denotes the causal temporal scanning. Averaging forward and backward scans eliminates directional bias while encoding precise spatial context, and the complete state-to-output formulation in~\cref{eq:recurrentST} recovers the enriched features $\mathbf{F}_{\text{ST}}[t,n] = \bar{\mathbf{C}}_t \mathbf{h}_{n,t}$ through the additional output matrix $\bar{\mathbf{C}}_t$, where the input-dependent matrices $\bar{\mathbf{A}}_t, \bar{\mathbf{B}}_t, \bar{\mathbf{C}}_t \in \mathbb{R}^{d_e \times d_e}$ jointly enable selective propagation while filtering redundancy. The SSA module consists of a single pair of spatial and temporal aggregation, each maintaining $d_e=1024$ dimensional hidden states with internal state dimension $d_{\text{state}}=16$ and expansion factor $2$. This scanning strategy is initialized randomly and trained from scratch, forming the core mechanism enabling generalization to reasoning tasks.

\para{Hierarchical Compression.} Following spatiotemporal enrichment with SSA, the Hierarchical State-adaptive Compression (HSC) module performs sequential token reduction through learned adaptive thresholds. Temporal compression first identifies motion-critical frames by computing frame-level similarity scores $s_t$ between consecutive frames. An adaptive threshold $\tau_{\text{temp}}$ determines candidates for merging based on similarity to their predecessor ($s_t > \tau_{\text{temp}}$). This produces a binary retention mask $\mathbf{m}_{\text{temp}} \in \{0,1\}^T$ where $\mathbf{m}_{\text{temp}}[t]=1$ indicates frame $t$ should be retained. Spatial compression then processes each temporally-retained frame independently, computing per-location coherence scores $c_n$ across the temporal dimension and adaptive thresholds $\tau_{\text{spatial}}^{(n)}$ to produce spatial retention masks $\mathbf{m}_{\text{spatial}}^{(t)} \in \{0,1\}^N$ for each frame.

The merging operation consolidates information into retained positions (where $\mathbf{m}[i]=1$). For each retained position $i$, all preceding consecutive non-retained positions are pooled together via averaging:
\begin{equation}
\begin{aligned}
\mathbf{F}_{\text{summary}}[i] &= \frac{1}{|\mathcal{G}_i|} \sum_{j \in \mathcal{G}_i} \mathbf{F}[j], \\
\text{where } \mathcal{G}_i &= \{j : i_{\text{prev}} < j \leq i, \mathbf{m}[j]=0 \text{ for } j \neq i\} \cup \{i\}.
\end{aligned}
\end{equation}
Here, $\mathcal{G}_i$ denotes the group of consecutive indices merged into position $i$. The combined compression and extraction with the mask $m$ yields the compressed features $\mathbf{F}_c \in \mathbb{R}^{T' \times N' \times d_e}$, that preserves motion-critical frames and dynamic spatial regions while consolidating static content. Further details on compression formulation and threshold learning are provided in the main paper.

\begin{table}[t]
\centering
\footnotesize 
\caption{Training configuration for STAC.}
\label{tab:training_config}
{%
\begin{tabular}{lc}
\toprule
\textbf{Component} & \textbf{Value} \\
\midrule
Context frame num & 2-32 \\
Query frame num & 4 \\
Input resolution & 224 \\
Feature downsampling rate & 4 \\
\midrule
Optimizer & AdamW \\
Optimizer momentum & $\beta_1, \beta_2 = 0.9, 0.999$ \\
Optimizer weight decay & 0.05 \\
Learning rate & $3 \times 10^{-4}$ \\
LoRA rank & 8 \\
LoRA $\alpha$ & 16 \\
\midrule
Gradient accumulation & 10 \\
Batch size (Per Device, Per Step) & 2 \\
Batch size (Overall) & 40 \\
Warmup steps & 100 \\
Total training steps & 3,000 \\
\midrule
Loss: $\lambda_{ce}$ & 1.0 \\
Loss: $\lambda_{dice}$ & 0.5 \\
Loss: $\lambda_{bce}$ & 2.0 \\
Loss: $\lambda_{comp}$ & 0.001 \\
\bottomrule
\end{tabular}%
}
\end{table}

\para{Multimodal LLM.} Compressed features pass through a vision-to-language projection layer $\phi_{V \rightarrow L}$ mapping from visual dimension $d_e=1024$ to language model dimension $D=4096$:
\begin{equation}
\mathbf{F}_{\text{LLM}} = \phi_{V \rightarrow L}(\mathbf{F}_c).
\end{equation}
The language model then processes these features auto-regressively to generate segmentation tokens. For each frame $t$ requiring segmentation, the MLLM receives both the compressed summary providing global context and the full-resolution spatiotemporally-enriched features for that frame:
\begin{equation}
\langle \text{SEG} \rangle_t = \text{LLM}([R, \phi_{V \rightarrow L}(\mathbf{F}_c), \phi_{V \rightarrow L}(\mathbf{F}_{\text{ST}}[t])]),
\end{equation}
where $R$ denotes the referring expression text. This hybrid strategy balances efficiency through compressed historical context with precision through full-resolution current frame features. 

\para{Mask Decoder.} Our mask decoder follows the design of SAM2~\cite{ravi2024sam}.  After obtaining $\langle \text{SEG} \rangle_t$, the token embedding $\hat{\mathbf{u}}_t$ is passed through a language-to-vision projection layer $\phi_{L \rightarrow V}$ to generate a prompt for the mask decoder. The mask decoder $\text{Dec}$ processes the prompt alongside encoded frame features from the vision encoder $\text{Enc}$, while maintaining a memory of features in the memory bank, to generate the segmentation mask.
\begin{equation}
M_t = \text{Dec}(\text{Enc}(\mathbf{I}_t), \phi_{L \rightarrow V}(\mathbf{u}_t), \text{FeatBank}).
\end{equation}

Here, the memory bank stores encoded features of previously generated masks and applies attention over up to 7 selected historical masks to provide temporal consistency. All decoder and memory components are jointly fine-tuned end-to-end with the STAC framework.

\para{Training Configuration.} 
We initialize the image encoder $\text{Enc}$, mask decoder $\text{Dec}$, memory encoder, and memory attention modules from pretrained SAM2 (sam2\_hiera\_large)~\cite{ravi2024sam} weights. The LLM backbone (LLaVA-7B~\cite{liu2023visual,touvron2023llama}), projection layers ($\phi_{V \rightarrow L}$, $\phi_{L \rightarrow V}$), and visual tokenizer (CLIP ViT-L/14~\cite{radford2021learning}) are initialized from available pretrained weights~\cite{lai2024lisa}. The LLM is fine-tuned via LoRA~\cite{hu2022lora} with rank $r=8$ and scaling $\alpha=16$ applied exclusively to attention projection matrices, with LoRA parameters initialized randomly. The training objective is given as:
\begin{equation}
\mathcal{L}_{\text{total}} = \lambda_{ce}\mathcal{L}_{ce} + \lambda_{dice}\mathcal{L}_{dice} + \lambda_{bce}\mathcal{L}_{bce} + \lambda_{comp}\mathcal{L}_{comp},
\end{equation}
where the training objective expands $\mathcal{L}_{\text{seg}}$ from the main paper into its constituent terms: cross-entropy loss ($\mathcal{L}_{ce}$), DICE loss ($\mathcal{L}_{dice}$), and per-pixel BCE loss ($\mathcal{L}_{bce}$), alongside the compression regularisation ($\mathcal{L}_{comp}$) with $\lambda_{comp}=0.001$. The training hyperparameters are summarized in ~\cref{tab:training_config}.

\begin{algorithm}[!t]
    \footnotesize
    \caption{STAC Offline Inference}
    \label{alg:offline}
    \begin{algorithmic}[1]
        \STATE \textbf{Input:} Video $\mathbf{V} = \{\mathbf{I}_1, \ldots, \mathbf{I}_T\}$, text query $q$
        \STATE \textbf{Output:} Segmentation masks $\mathbb{M} = \{m_1, \ldots, m_T\}$
        
        \STATE $\mathbf{F}_e \gets \text{Encoder}(\mathbf{V})$ 
        \STATE \textbf{// Stage 1: State-informed Spatiotemporal Aggregator (SSA)}
        \STATE \text{// Spatial Aggregation}
        \FOR{$t = 1$ to $T$ \textbf{in parallel}}
            \STATE $\mathbf{F}_{\text{spatial}}[t] \gets \frac{1}{2}\left(\mathcal{M}^b_{\rightarrow}(\mathbf{F}_e[t]) + \mathcal{M}^b_{\leftarrow}(\mathbf{F}_e[t])\right)$
        \ENDFOR
        \STATE \text{// Temporal Aggregation}
        \FOR{$n = 1$ to $N$ \textbf{in parallel}}
            \STATE $\mathbf{h}_{n,t} \gets \bar{\mathbf{A}}_t \mathbf{h}_{n,t-1} + \bar{\mathbf{B}}_t \mathbf{F}_{\text{spatial}}[t,n], \quad t=1,\ldots,T$
            \STATE $\mathbf{F}_{\text{ST}}[:,n] \gets \mathcal{M}_{\text{causal}}(\mathbf{F}_{\text{spatial}}[:,n])$
            \STATE Store hidden state: $\mathbf{h}_n^* \gets \mathbf{h}_{n,T}$
        \ENDFOR
        
        \STATE \textbf{// Stage 2: Adaptive Hierarchical Compression}
        \STATE $\tau_{\text{temp}} \gets \text{AdaptiveThreshold}(\mathbf{F}_{\text{ST}}, \text{temporal})$
        \STATE $\mathbf{F}_{\text{temp}} \gets \text{Compress}_{\text{temporal}}(\mathbf{F}_{\text{ST}}, \tau_{\text{temp}}) \quad \text{// } T \rightarrow T'$
        \STATE $\{\tau_{\text{spatial}}^{(n)}\}_{n=1}^N \gets \text{AdaptiveThreshold}(\mathbf{F}_{\text{temp}}, \text{spatial})$
        \STATE $\mathbf{F}_c \gets \text{Compress}_{\text{spatial}}(\mathbf{F}_{\text{temp}}, \{\tau_{\text{spatial}}^{(n)}\}) \quad \text{// } N \rightarrow N'$
        \STATE \text{// Last Frame is always kept with full feats.}
        
        \STATE \textbf{// Stage 3: Frame-by-Frame Segmentation}
        \STATE $\mathbf{F}_{\text{summary}} \gets \text{Proj}(\mathbf{F}_c)$
        \FOR{$t = 1$ to $T$}
            \STATE $\mathbf{F}_{\text{full}}^{(t)} \gets \text{SSA}(\mathbf{I}_t, \{\mathbf{h}_n^*\}_{n=1}^N) \quad \text{// Enrich Frame}$
            \STATE $\text{trk}_t \gets \text{MLLM}(q, \mathbf{F}_{\text{summary}}, \text{Proj}(\mathbf{F}_{\text{full}}^{(t)}))$
            \STATE $m_t \gets \text{Mask-Predictor}(\text{trk}_t, \mathbf{I}_t)$
            
        \ENDFOR
        \STATE \textbf{return} $\{m_t\}_{t=1}^T$ 
    \end{algorithmic}
\end{algorithm}

\begin{algorithm}[!t]
    \footnotesize
    \caption{STAC Online Inference}
    \label{alg:online}
    \begin{algorithmic}[1]
        \STATE \textbf{Input:} Video stream $\{\mathbf{I}_1, \mathbf{I}_2, \ldots\}$, text query $q$
        \STATE \textbf{Output:} Masks $\{m_1, m_2, \ldots\}$ (incremental)
        
        \STATE \textbf{// Initialize}
        \STATE $\mathbf{F}_{\text{summary}} \gets \emptyset$, $\{\mathbf{h}_{n,0}\}_{n=1}^N \gets 0$, $t \gets 0$
        \STATE $\{\mu, \sigma\}_{\text{temp/spatial}} \gets 0 \quad \text{// Running statistics}$
        
        \WHILE{new frame $\mathbf{I}_t$ arrives}
            \STATE $t \gets t + 1$
            
            \STATE $\mathbf{F}_e[t] \gets \text{Encoder}(\mathbf{I}_t)$ 
            \STATE \textbf{// Stage 1: Incremental SSA}
            \STATE \text{// Spatial Aggregation (frame-independent)}
            \STATE $\mathbf{F}_{\text{spatial}}[t] \gets \frac{1}{2}\left(\mathcal{M}^b_{\rightarrow}(\mathbf{F}_e[t]) + \mathcal{M}^b_{\leftarrow}(\mathbf{F}_e[t])\right)$
            \STATE \text{// Temporal Aggregation (update hidden states)}
            \FOR{$n = 1$ to $N$ \textbf{in parallel}}
                \STATE $\mathbf{h}_{n,t} \gets \bar{\mathbf{A}}_t \mathbf{h}_{n,t-1} + \bar{\mathbf{B}}_t \mathbf{F}_{\text{spatial}}[t,n]$
                \STATE $\mathbf{F}_{\text{ST}}[t,n] \gets \mathcal{M}_{\text{causal}}(\mathbf{F}_{\text{spatial}}[t,n], \mathbf{h}_{n,t})$
            \ENDFOR
            
            \STATE \textbf{// Stage 2: Adaptive Compression}
            \STATE Update $\{\mu, \sigma\}_{\text{temp}}$ \text{// with frame $t$}
            \STATE $\tau_{\text{temp}}^{(t)} \gets \text{AdaptiveThreshold}(\{\mu, \sigma\}_{\text{temp}})$
            \STATE $\mathbf{F}_{\text{temp}}[t] \gets \text{Compress}_{\text{temporal}}(\mathbf{F}_{\text{ST}}[t], \tau_{\text{temp}}^{(t)})$
            \STATE Update $\{\mu, \sigma\}_{\text{spatial}}$ \text{// with frame $t$}
            \STATE $\{\tau_{\text{spatial}}^{(n)}\}^{(t)} \gets \text{AdaptiveThreshold}(\{\mu_c^{(n)}, \sigma_c^{(n)}\})$
            \STATE $\mathbf{F}_c[t] \gets \text{Compress}_{\text{spatial}}(\mathbf{F}_{\text{temp}}[t], \{\tau_{\text{spatial}}^{(n)}\}^{(t)})$
            \STATE $\mathbf{F}_{\text{summary}} \gets \mathbf{F}_{\text{summary}} \oplus \mathbf{F}_c[t]$
            
            \STATE \textbf{// Stage 3: Generate Segmentation Mask}
            \STATE $\text{trk}_t \gets \text{MLLM}(q, \text{Proj}(\mathbf{F}_{\text{summary}}), \text{Proj}(\mathbf{F}_{\text{ST}}[t]))$
            \STATE $m_t \gets \text{Mask-Predictor}(\text{trk}_t, \mathbf{I}_t)$
            \STATE \textbf{output} $m_t$
        \ENDWHILE
    \end{algorithmic}
\end{algorithm}

\subsection{Inference Procedures}

STAC supports two inference modes designed for different deployment scenarios.

\para{Offline Batch Processing.} When complete video access is available, STAC processes all $T$ frames through the SSA module to enrich each token with comprehensive spatiotemporal context. The hierarchical compression module then operates sequentially to perform a temporal-then-spatiotemporal compression, producing a compact summary consisting of $T'N'$ tokens.

For mask generation, STAC adopts a hybrid strategy that balances global efficiency with local precision. The compressed summary $\mathbf{F}_c$ representing entire video context provides global understanding to the MLLM. For each frame $I_t$ requiring segmentation,the architecture is designed to provide full-resolution spatiotemporally-enriched features $\mathbf{F}_{\text{ST}}[t]$ containing all $N$ spatial tokens for that frame. The MLLM processes both $[\phi_{V \rightarrow L}(\mathbf{F}_c), \phi_{V \rightarrow L}(\mathbf{F}_{\text{ST}}[t])]$, enabling compressed historical context for temporal reasoning while maintaining complete spatial detail for precise boundary localization. This prevents loss of fine-grained spatial information that would occur if segmentation relied solely on compressed representation. The complete offline procedure is detailed in ~\cref{alg:offline}.

\para{Online Streaming.} Real-time scenarios where frames arrive sequentially require incremental processing without future frame access. Unlike offline mode which processes all frames simultaneously, online mode maintains hidden states $\{\mathbf{h}_{n,t}\}_{n=1}^{N}$ in the SSA for each of $N$ spatial locations, accumulating temporal context as frames arrive. When frame $I_t$ becomes available, it first passes through spatial aggregation with frame independent bidirectional scanning. The spatially-aggregated features then enter the causal temporal scan, which leverages existing hidden states to produce a well-informed representation while simultaneously updating these states with current information via the recurrent formulation in~\cref{eq:recurrentST}, processed in parallel across all spatial locations $n$. This avoids recomputing past representations, enabling true incremental processing with constant per-frame computational cost.

Adaptive thresholds in the HSC module evolve dynamically during online inference. Rather than computing statistics over a complete video as in offline mode, the network maintains running estimates of mean $\mu_t$ and standard deviation $\sigma_t$ over observed frames $[1, t]$. These running statistics feed threshold prediction MLPs to generate adaptive thresholds at each timestep. Following compression, the compressed features are concatenated with full-resolution features of the current frame and passed through the MLLM to generate the mask prediction. The complete online procedure is detailed in~\cref{alg:online}

\section{Additional Ablation Studies}

\label{sec:supp_ablations}

\begin{figure}[t]
    \centering
    \includegraphics[width=0.95\columnwidth]{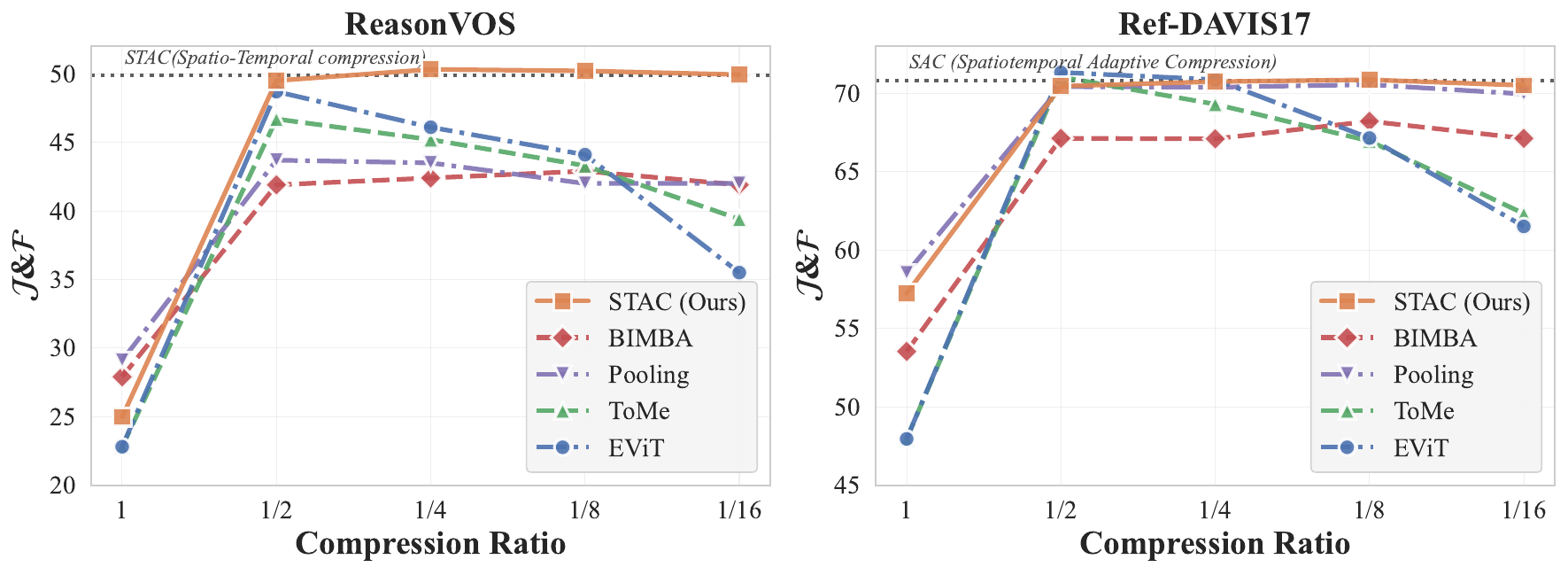}
    \caption{Spatial compression strategy comparison shows that STAC maintains robust performance across compression ratios by preserving motion-critical tokens identified through SSA.}
    \label{fig:spatial_compression}
\end{figure}

\begin{table}[t]
\footnotesize
\centering
\caption{Effect of local query frame multiplicity on ReasonVOS. }
\label{tab:query_frames}
\resizebox{0.5\columnwidth}{!}{%
\begin{tabular}{c|ccc|ccc}
\toprule
\multirow{2}{*}{\textbf{Local Queries}} & \multicolumn{3}{c|}{\textbf{GLUS}} & \multicolumn{3}{c}{\textbf{STAC (Ours)}} \\
\cmidrule(lr){2-4} \cmidrule(lr){5-7}
& $\mathcal{J}$\&$\mathcal{F}$ & $\mathcal{J}$ & $\mathcal{F}$ & $\mathcal{J}$\&$\mathcal{F}$ & $\mathcal{J}$ & $\mathcal{F}$ \\
\midrule
1 & 48.1 & 45.4 & 51.0 & \textbf{49.6} & \textbf{47.2} & \textbf{52.0} \\
2 & 45.6 & 42.7 & 48.5 & 48.9 & 46.5 & 51.3 \\
3 & 44.3 & 41.6 & 47.0 & 49.1 & 46.8 & 51.4 \\
4 & 47.2 & 44.4 & 50.0 & 48.5 & 46.3 & 50.7 \\
\bottomrule
\end{tabular}%
}
\end{table}

\subsection{Spatial Compression Strategy Comparison} 
\label{ssec:spatial_compression}

We evaluate whether our coherence-based compression effectively preserves the spatiotemporal representations learned by SSA scanning compared to alternative token selection strategies. The STAC, BIMBA~\cite{islam2025bimba}, and Pooling are trained end-to-end with their respective compression modules on the same backbone, while ToMe~\cite{bolya2022token} and EViT~\cite{liang2022not} are applied on the trained GLUS~\cite{lin2025glus} model. We then vary compression ratios $r \in \{1, 1/2, 1/4, 1/8, 1/16\}$ at inference and evaluate on ReasonVOS~\cite{bai2024one} and Ref-DAVIS17~\cite{khoreva2018video}. As shown in Figure~\ref{fig:spatial_compression}, STAC maintains remarkably stable performance across all ratios, exhibiting minimal degradation even at extreme 16$\times$ reduction. The competing methods, however, suffer substantial performance collapse under aggressive compression, with attention-based approaches (ToMe, EViT) degrading most severely. This disparity arises because attention-based methods prioritize within-frame saliency while discarding the temporal context they never explicitly modeled. In contrast, our SSA scanning mechanism encodes motion-critical information into spatiotemporally enriched representations that prove inherently robust to aggressive token reduction. Our coherence-based selection complements this robustness by explicitly identifying dynamic regions through cross-frame consistency, preserving key details learned from the scanning strategy.

\begin{figure*}[t!]
    \centering
    \includegraphics[width=0.85\textwidth]{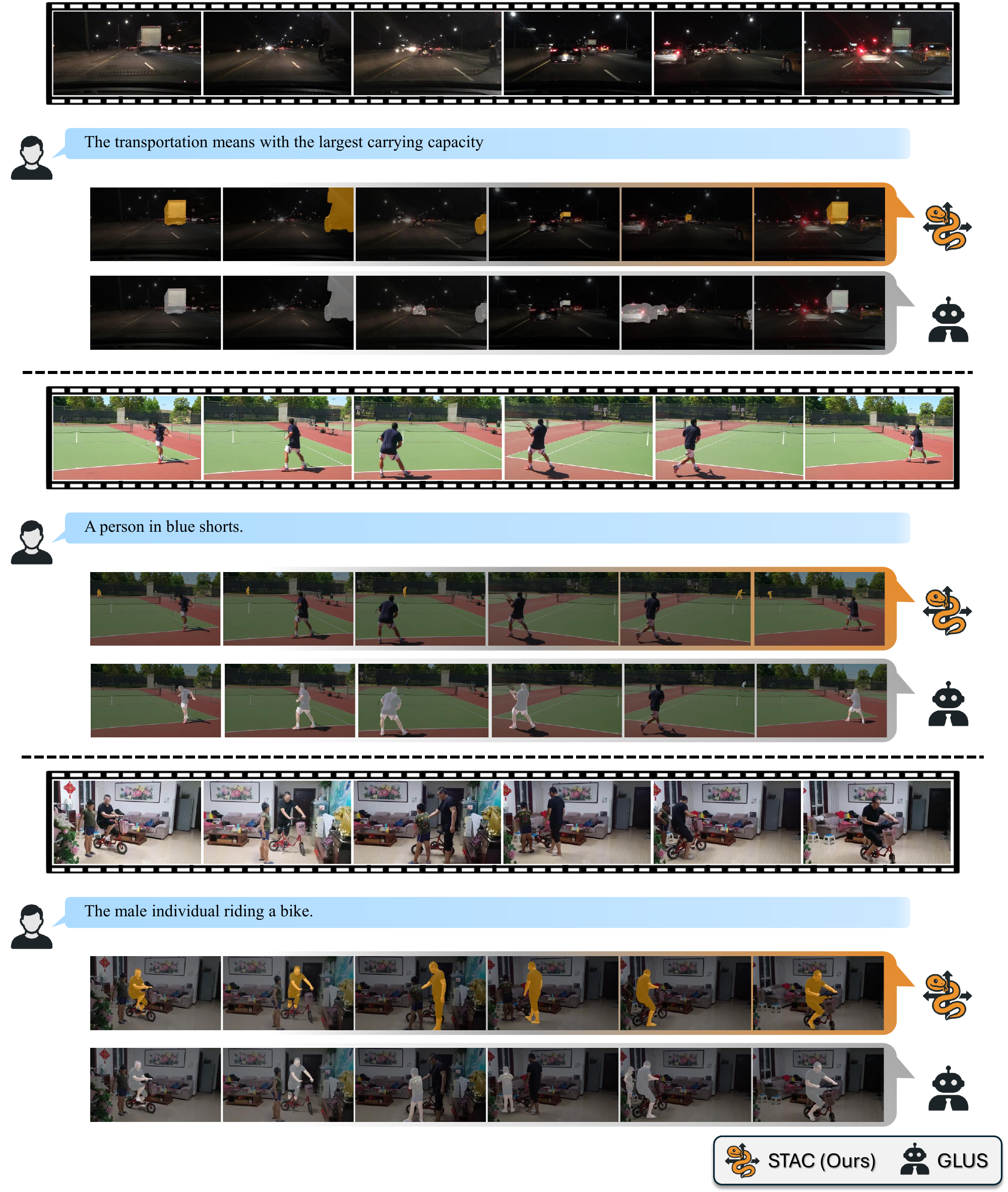}
    \caption{\textbf{Temporal Consistency and Object Identity Preservation.} STAC maintains stable segmentation of a target object across frames despite partial occlusions and viewpoint changes. The causal temporal scanning propagates object identity through selective state updates, producing smooth mask transitions without flickering.}
    \label{fig:successviz1}
    \vspace{-1.5em}
\end{figure*}

\subsection{Impact of Local Query Frame}
\label{ssec:query_frames}

We examine whether providing multiple consecutive full-resolution local query frames enhances segmentation by supplying additional local detail beyond the compressed global context. For fair comparison, we train both STAC and the baseline GLUS~\cite{lin2025glus} using identical settings with a fixed context length. At inference, we vary query frames from $Q=1$ to $Q \in \{2,3,4\}$ and evaluate on the ReasonVOS~\cite{bai2024one} dataset. As observed in Table~\ref{tab:query_frames}, STAC maintains remarkably stable performance with minimal variation (within 1.1 $\mathcal{J}$\&$\mathcal{F}$ points) across all configurations while consistently outperforming the baseline. This stability demonstrates that the SSA scanning strategy successfully encodes motion-critical information into spatiotemporally enriched representations, rendering additional local frames redundant and supporting deployment efficiency where a single query frame suffices for accurate segmentation.

\begin{figure*}[t!]
    \centering
    \includegraphics[width=0.85\textwidth]{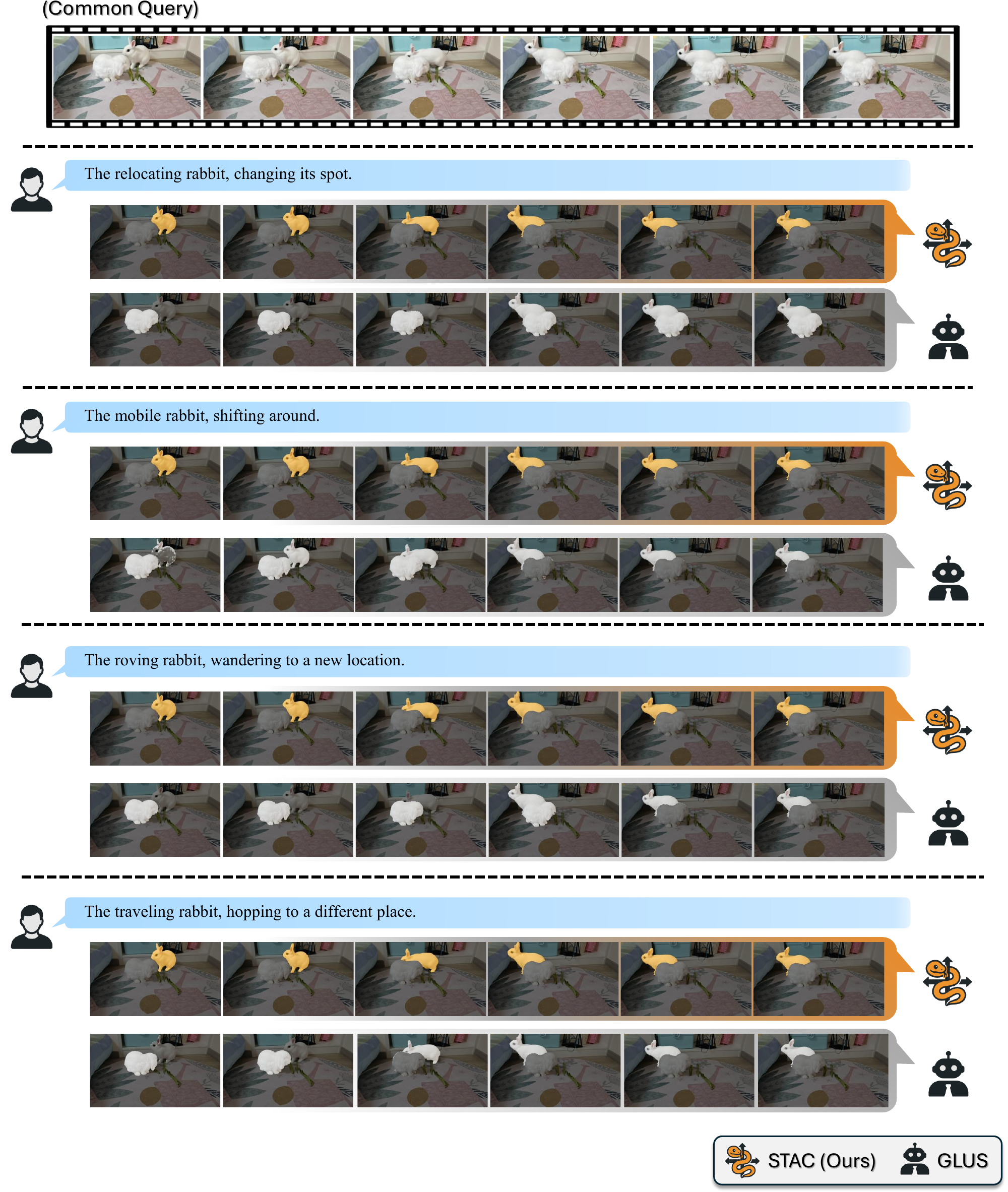}
    \caption{\textbf{Linguistic Invariance and Cross-Query Consistency.} Given multiple semantically equivalent but lexically distinct queries referring to the same object, STAC produces identical segmentation masks. This demonstrates that the dual-stage scanning learns object-centric representations invariant to linguistic variation.}
    \label{fig:successviz2}
\end{figure*}

\begin{figure*}[t!]
    \centering
    \includegraphics[width=0.815\textwidth]{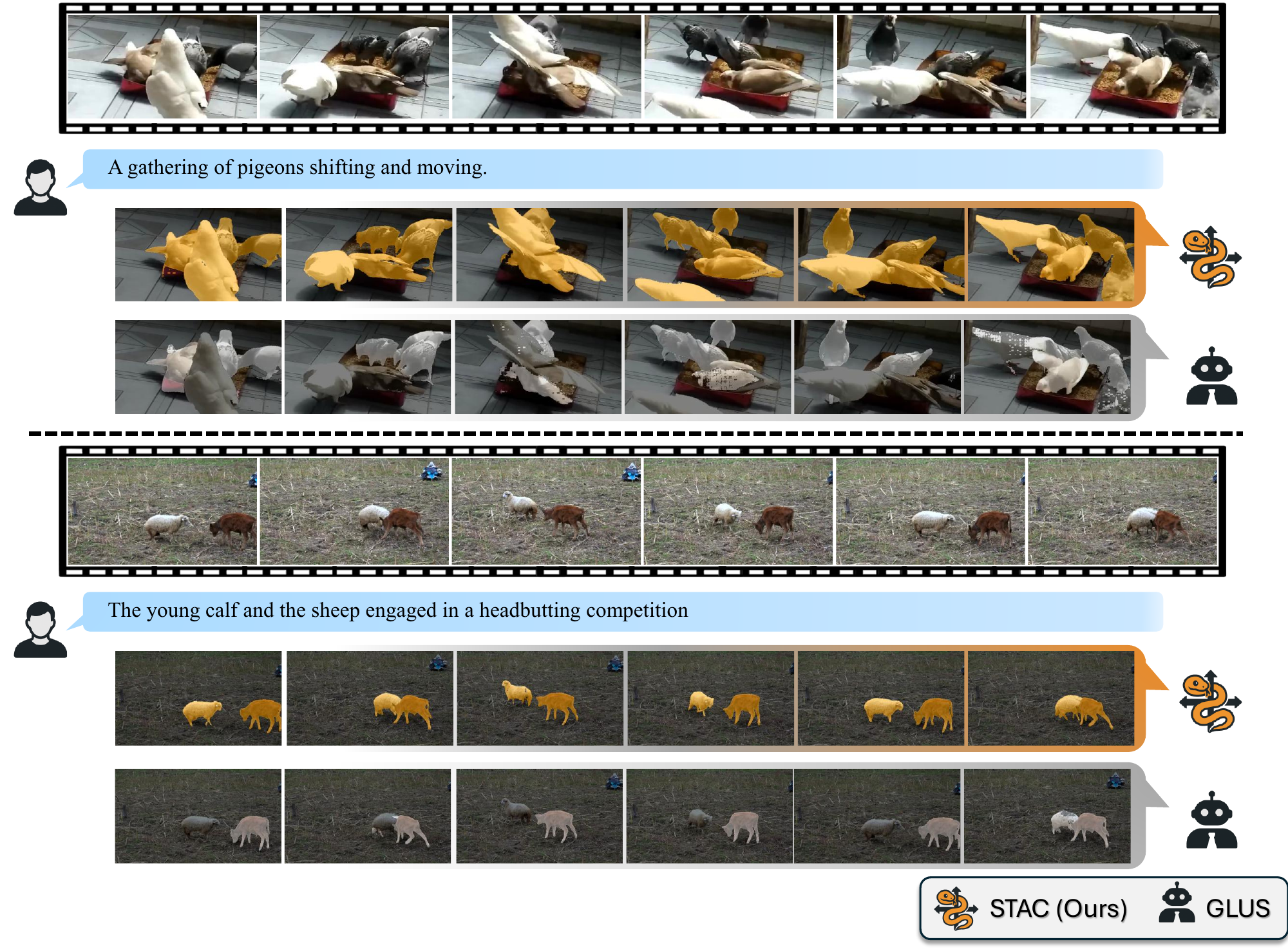}
    \vspace{1px}
    \caption{\textbf{Complex Multi-Object Reasoning.} STAC correctly identifies and segments multiple referenced objects within a single forward pass while maintaining distinct instance boundaries and temporal consistency for each target throughout the sequence.}
    \label{fig:successviz3}
\end{figure*}
\vspace{1px}

\begin{figure*}[t!]
    \centering
    \includegraphics[width=0.83\textwidth]{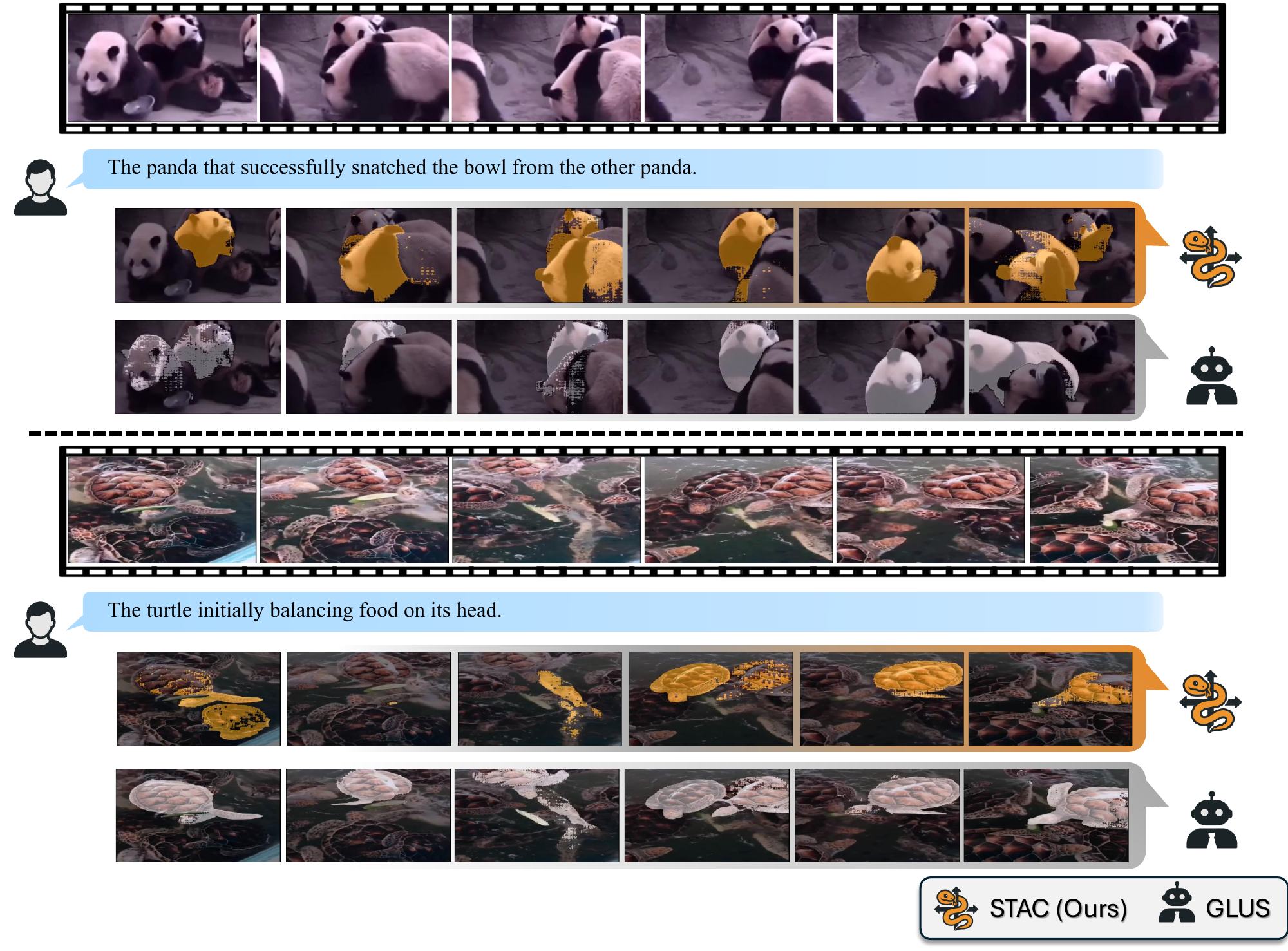}
    \vspace{1px}
    \caption{\textbf{Challenging Scenarios.} Scenes with multiple visually similar objects exhibiting complex simultaneous motions present difficulties for video reasoning systems. Subtle boundary distinctions may blur when spatially coherent regions are merged during compression.}
    \label{fig:complexviz}
\end{figure*}

\section{Extended Qualitative Analysis}
\label{sec:supp_qualitative}

\subsection{Success Cases}
\label{ssec:success_cases}

The following examples demonstrate the proposed STAC architecture's ability to maintain temporal consistency and spatial precision through its dual-stage scanning strategy in challenging video reasoning scenarios.

\para{Temporal Consistency and Object Identity Preservation.} STAC maintains stable segmentation across frames when provided with unambiguous referring expressions. As observed in \cref{fig:successviz1}, the model demonstrates consistent object tracking even under partial occlusion and viewpoint changes, producing smooth frame transitions with minimal mask flickering. This stability stems from our causal temporal scanning, which propagates object identity through selective hidden state updates while filtering temporal noise. The hierarchical compression preserves these temporally coherent representations despite aggressive reduction.

\para{Linguistic Invariance and Cross-Query Consistency.} When presented with multiple distinct but unambiguous queries referring to the same object (\cref{fig:successviz2}), STAC consistently segments the identical target across all linguistic variations. This demonstrates that our dual-stage scanning learns spatiotemporally enriched representations capturing object-centric features invariant to phrasing differences. The use of bidirectional spatial scanning encodes discriminative appearance features while causal temporal scanning maintains object identity, enabling reliable cross-query identification.

\para{Complex Multi-Object Reasoning.} STAC successfully handles queries referring to multiple objects simultaneously, correctly identifying and segmenting all referenced targets in the scene with a single forward pass while maintaining distinct boundaries. As shown in \cref{fig:successviz3}, the model preserves temporal consistency for each object individually, demonstrating its capacity to reason about complex multi-object spatiotemporal relationships. This capability emerges from the scanning strategy's ability to learn rich interactions between multiple dynamic entities, with hierarchical compression preserving motion-critical regions where objects interact.

\subsection{Challenging Scenarios}
\label{ssec:challenging_scenarios}

While STAC demonstrates strong performance across diverse benchmarks, scenes involving multiple visually similar objects with complex simultaneous motions remain challenging. As shown in \cref{fig:complexviz}, such scenarios require joint reasoning about fine-grained appearance differences, spatial relationships, and individual motion patterns across extended sequences. Our compression framework, which prioritizes efficiency by merging spatially coherent regions, may occasionally blur subtle boundaries between similar adjacent objects. Notably, baseline methods without compression exhibit comparable difficulties, suggesting this reflects fundamental limitations of current video reasoning architectures rather than compression-specific artifacts. These cases represent opportunities for future work on fine-grained discrimination mechanisms.

\subsection{Video of Qualitative Results}

A supplementary video accompanies this submission, visualizing both segmentation results and compression dynamics across sequences. To illustrate compression behavior, we employ a red gradient overlay where increasing intensity indicates patches undergoing sequential aggregation, with color refresh marking summary updates of accumulated information.

\section{Future Directions}
\label{ssec:future_directions}

Our work primarily focuses on designing efficient spatiotemporal scanning strategies for video reasoning segmentation, introducing a novel dual-stage mechanism that successfully learns generalizable representations despite training only on referring data. Building upon this foundation, meaningful future work could explore specialized mechanisms for fine-grained appearance discrimination in scenarios involving multiple visually similar objects, potentially through attention-based feature refinement that operates on the rich spatiotemporal representations learned by our scanning strategy while preserving computational efficiency. Additionally, hierarchical memory structures operating at multiple temporal scales could further enhance the model's ability to maintain fine-grained distinctions across extended sequences. 

The success of our dual-stage scanning principle suggests promising extensions to broader video understanding domains including audio-visual reasoning and embodied AI, where the architectural separation of spatial and temporal processing provides a general principle for efficient spatiotemporal reasoning. We hope that combining STAC's scanning architecture with longer context windows and video-native pre-training could further unleash its potential across these diverse application domains.

\vspace{1px}

\newpage

%
%
\bibliographystyle{splncs04}
\bibliography{main}